\documentclass[sigconf]{acmart} 
\AtBeginDocument{%
  \providecommand\BibTeX{{%
    \normalfont B\kern-0.5em{\scshape i\kern-0.25em b}\kern-0.8em\TeX}}}

\usepackage{microtype}
\usepackage{graphicx}
\usepackage{booktabs}
\usepackage{multirow}

\usepackage{amsmath}
\usepackage{mathtools}
\usepackage{amsthm}
\usepackage{bm}
\usepackage{amsfonts}       
\usepackage{nicefrac}       
\usepackage{algorithm}
\usepackage{algorithmic}
\usepackage{subcaption}

\copyrightyear{2022}
\acmYear{2022}
\setcopyright{rightsretained}
\acmConference[BCB '22]{13th ACM International Conference on Bioinformatics, Computational Biology and Health Informatics}{August 7--10, 2022}{Northbrook, IL, USA}
\acmBooktitle{13th ACM International Conference on Bioinformatics, Computational Biology and Health Informatics (BCB '22), August 7--10, 2022, Northbrook, IL, USA}
\acmDOI{10.1145/3535508.3545563}
\acmISBN{978-1-4503-9386-7/22/08}

\begin{document}

\title{EvoVGM: a Deep Variational Generative Model for Evolutionary Parameter Estimation}


\author{Amine M. Remita}
\email{remita.amine@courrier.uqam.ca}
\orcid{0000-0003-4471-3195}
\affiliation{%
  \institution{Department of Computer Science\\ Universit\'e du Qu\'ebec \`a Montr\'eal}
  \streetaddress{P.O. Box 8888 Downtown Station}
  \city{Montreal}
  \state{Quebec}
  \postcode{H3C 3P8}
  \country{Canada}
}

\author{Abdoulaye Banir\'e Diallo}
\orcid{0000-0002-1168-9371}
\email{diallo.abdoulaye@uqam.ca}
\affiliation{%
  \institution{Department of Computer Science\\ Universit\'e du Qu\'ebec \`a Montr\'eal}
  \streetaddress{P.O. Box 8888 Downtown Station}
  \city{Montreal}
  \state{Quebec}
  \postcode{H3C 3P8}
  \country{Canada}
}

\renewcommand{\shortauthors}{Remita and Diallo}

\begin{abstract}
Most evolutionary-oriented deep generative models do not explicitly consider the underlying evolutionary dynamics of biological sequences as it is performed within the Bayesian phylogenetic inference framework.
In this study, we propose a method for a deep variational Bayesian generative model (EvoVGM) that jointly approximates the true posterior of local evolutionary parameters and generates sequence alignments.
Moreover, it is instantiated and tuned for continuous-time Markov chain substitution models such as JC69, K80 and GTR. 
We train the model via a low-variance stochastic estimator and a gradient ascent algorithm.
Here, we analyze the consistency and effectiveness of EvoVGM on synthetic sequence alignments simulated with several evolutionary scenarios and different sizes. Finally, we highlight the robustness of a fine-tuned EvoVGM model using a sequence alignment of gene S of coronaviruses.
\end{abstract}




\maketitle

\section{Introduction}
In systematics and evolutionary biology, probabilistic evolutionary models are extensively used to study unseen and complex historical events affecting the genomes of a set of taxa during a period of time (i.e., recombination, horizontal gene transfer and selective pressure).
Their ability to detect evolutionary events and measure their parameters using biological sequences has enabled valuable applications in population genetics \citep{kern2010}, medicine \citep{yuan2015} and epidemiology \citep{faria2014, dudas2017}.
These models allow the estimation of probabilities of certain types of mutations such as substitutions \citep{jc1969, tavare1986}, indels \citep{diallo2007} and genome rearrangements \citep{sankoff1999}. 
Main approaches supporting evolutionary studies, such as phylogenetics, implement evolutionary models with Markovian properties \citep{tavare1986}.

Typically, evolutionary parameters of these models are jointly represented with different types of high-dimensional variables (discrete and continuous), inducing a computationally intractable joint posterior.
Bayesian phylogenetic approaches provide methods to efficiently approximate the intractable joint posterior and quantify the uncertainty in the estimation of the parameters \citep{Yang1997, Huelsenbeck2001mrbayes}. They mainly implement random-walk Markov Chain Monte Carlo (MCMC) algorithms, which can converge to an accurate posterior but with a considerable cost. Furthermore, they are prone to limitations due to the complexity of the posterior \citep{whidden2015}, their dependence on initialization and proposal distribution parameters, and their sensitivity to the prior distributions \citep{Huelsenbeck2002}.
Recently, variational inference (VI) has sparked interest in phylogenetics as a robust alternative to approximate the intractable posterior by relying on fast optimization methods \citep{Dang2019, Fourment2019, Zhang2019, Zhang2020}.
VI finds an optimal candidate from a space of tractable distributions that minimizes the Kullback-Leibler (KL) divergence to the exact posterior \cite{Jordan1999, Blei2017}. It inherently bounds the intractable marginal likelihood of the observed data.
Moreover, VI is also used in building deep generative models \citep{Kingma2014, Rezende2014}.
However, contrary to Bayesian phylogenetic inference frameworks, most evolutionary-oriented deep generative models do not explicitly consider the underlying evolutionary dynamics of the biological sequences \citep{Riesselman2018, Lim2020, Weinstein2021}.

Here, we propose EvoVGM, a deep variational generative model that simultaneously estimates local evolutionary parameters and generates nucleotide sequence data. Like phylogenetic inference, we explicitly integrate a continuous-time Markov chain substitution model into the generative model. The model is trained in an unsupervised manner following the evolutionary model constraints.

\section{Background}
\subsection{Notation}
The observed data \(\mathbf{X}\) is an alignment of \(M\) character sequences with length \(N\), where \(\mathbf{X} \in \mathcal{A}^{M \times N}\). In our case, the alphabet of characters \(\mathcal{A} = \{A, G, C, T\}\) is a set of nucleotides. 
\(x_n^m\) is the character in the \(m\)\textsuperscript{th} sequence (\(x^m\)) and at the \(n\)\textsuperscript{th} site (\(x_n\)) of the alignment.
Here, we assume that each alignment \(\mathbf{X}\) has a hidden ancestral state sequence \(a \in \mathcal{A}^{N}\). We take the hypothesis that each ancestral state \(a_n\) has evolved independently from the other states \(\{ a_i; i \neq n\}\) to an extant character \(x_n^m\) over an evolutionary time expressed as a branch length \(t\) and following a substitution model defined by a set of parameters \(\psi\).
In a Bayesian framework, we seek representations allowing to model uncertainty on the quantity and the composition of different entities. We consider the observable characters (\(\mathbf{x}_n^m\)) and the ancestral states (\(\mathbf{a}_n\)) as random variables (noted in bold, unlike scalar values) and represent them by categorical distributions over \(\mathcal{A}\). Also, branch lengths (\(\mathbf{t}^m\)) and substitution model parameters (\(\bm{\psi}\)) will be modelled as random variables and will be represented by suitable distributions.

\subsection{Markov Chain Models of Character Substitution}
The evolution of a character is measured by the number of hidden substitutions that undergoes over time. To estimate this quantity, we assume that the process of evolution follows a continuous-time Markov chain model whose states belong to \(\mathcal{A}\). The model is parameterized by a rate matrix \(\mathbf{Q}\) and relative frequencies \(\pi\) of characters at equilibrium. Each element of the matrix \(q_{ij}\) (\( i\neq j\)) defines the instantaneous substitution rate of character \(i\) changing into character \(j\). The diagonal elements \(q_{ii}\) are set up in a way that each row sums to 0. \(\mathbf{Q}\) is scaled by a factor \(\mu\), so that the time \(\textit{t}\) will be measured in the expected number of substitutions per site and the average rate of substitution at equilibrium will be 1. We use time-reversible Markov chain models assuming the amount of changes from one character to another is the same in both ways. For nucleotide substitution time-reversible models, the equation of \(\mathbf{Q}\) is
\[
    \mathbf{Q} = \begin{pmatrix} \cdot &  a\pi_G &  b\pi_C &  c\pi_T \\
                                a\pi_A &  \cdot  &  d\pi_C &  e\pi_T \\
                                b\pi_A &  d\pi_G &  \cdot  &  f\pi_T \\
                                c\pi_A &  e\pi_G &  f\pi_C &  \cdot      
                \end{pmatrix} \mu,
\]
where \(a, b, c, d, e, \text{and} f\) are the set of relative substitution rate parameters \(\rho\), and \(\pi_A + \pi_G + \pi_C + \pi_T = 1\) are the relative frequencies \(\pi\). Once \(\mathbf{Q}\) is estimated we can compute the probability transition matrix \(\mathbf{P}\) over an evolutionary time \(\textit{t}\) as 
\(\mathbf{P}(\textit{t}) = \exp(\mathbf{Q}\,\textit{t}).\)
The matrix exponential is computed using spectral decomposition of \(\mathbf{Q}\) as it is reversible
(see \citep{lemey2009phylogenetic} and \citep{yang2014molecular} for more details).

Several substitution models could be generated depending on the constraints placed on the set of parameters \(\psi = \{\rho\), \(\pi\}\). The simplest model is JC69 with equal substitution rates and uniform relative frequencies \citep{jc1969}. The K80 model defines uniform frequencies like JC69, but it differentiates between the two types of substitution rates corresponding to transitions (\(\alpha = a = f\)) and transversions \(\beta = b = c = d = e\) \citep{kimura1980}. Usually, K80 is parameterized by the transition/transversion rate ratio \(\kappa = \alpha/\beta\). 
Finally, the general time-reversible (GTR) model sets all the parameters \(\psi\) free \citep{tavare1986,yang1994}.

\subsection{Evolutionary Posterior}
Along with \(\mathbf{a}\) and \(\mathbf{t}\) variables, we consider the parameters of the Markov chain model \(\bm{\psi}\) as latent (hidden) variables to be inferred from the observed data \(\mathbf{X}\).
Assuming an independent evolution of the sites in an alignment \citep{felsenstein1981}, the marginal likelihood of the data \(\mathbf{X}\) factorizes into \(p(\mathbf{X}) = \prod_{n=1}^{N} p(\mathbf{x}_n)\). The inference of the latent variables for each site \(x_n\) requires the computation of the evolutionary joint posterior \(p(\mathbf{a}_n, \mathbf{t}, \bm{\psi} | \mathbf{x}_n)\). The evolutionary posterior is calculated according to Bayes' theorem: 
\begin{equation}
    \label{bayes}
    p(\mathbf{a}_n, \mathbf{t}, \bm{\psi} | \mathbf{x}_n) = \frac{p(\mathbf{x}_n, \mathbf{a}_n, \mathbf{t}, \bm{\psi})}{p(\mathbf{x}_n)},
\end{equation}
which exposes the joint density of the observable variable and the latent variables \(p(\mathbf{x}_n, \mathbf{a}_n, \mathbf{t}, \bm{\psi})\), and the marginal likelihood \(p(\mathbf{x}_n)\). 
The former is factorized as a product of the joint prior density of the latent variables \(p(\mathbf{a}_n, \mathbf{t}, \bm{\psi})\) and the likelihood \(p(\mathbf{x}_n | \mathbf{a}_n, \mathbf{t}, \bm{\psi})\).
The latter is calculated by marginalizing over the values of all the latent variables as \(\iiint p(\mathbf{a}_n, \mathbf{t}, \bm{\psi})\, p(\mathbf{x}_n | \mathbf{a}_n, \mathbf{t}, \bm{\psi})\, d\mathbf{a}_n\, d\mathbf{t}\, d\bm{\psi}\). 
The computation of the evolutionary joint posterior density is computationally intractable as it depends on the evaluation of \( p(\mathbf{x}_n)\), which is intractable due to the integrals in its marginalization. 
We show in the next section strategies to determine each term in the equation \ref{bayes}.
\section{Proposed Evolutionary Model}
\label{methods}

\begin{figure}[b]
    \centering
    \includegraphics[width=0.9\columnwidth]{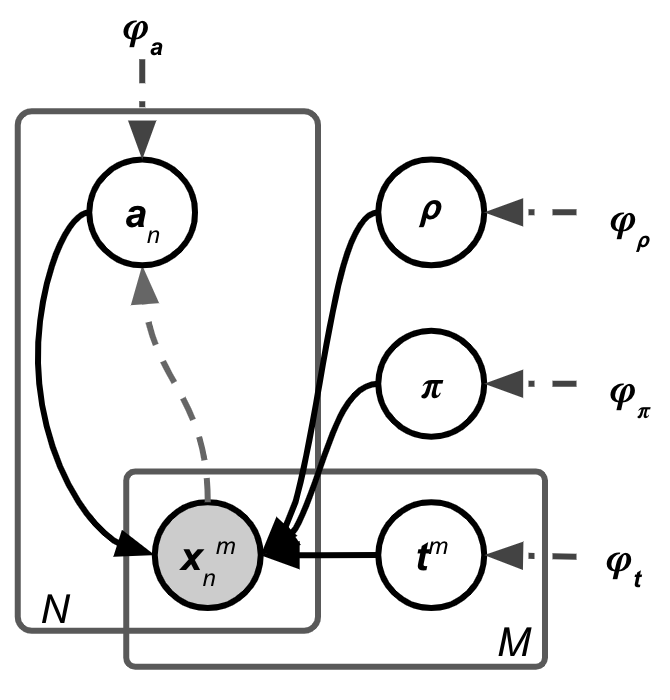} 
    \caption{Graphical illustration of the inference (dashed gray lines) and the generation (solid lines) processes of the GTR-based variational generative model.
    Gray circles represent the observed variables. Blank circles represent the latent variables.
    \(\{\varphi_{\mathbf{a}}, \varphi_{\mathbf{t}}, \varphi_{\bm{\rho}}, \varphi_{\bm{\pi}}\}\) is the set of hyper-parameters of the prior densities.}
    \label{fig:plate}
\end{figure}

In this section, we describe a deep variational generative model that simultaneously estimates local evolutionary biological parameters and generates nucleotide sequence data.
Similar to deep variational-based generative models \citep{Kingma2014,Rezende2014}, the proposed model architecture consists of two main sub-models: 1) a set of deep variational encoders that infers the parameters of evolutionary-latent-variable distributions and allows sampling, and 2) a generating model that computes probability transition matrices from sampled latent variables and generates a distribution of sequence alignments from reconstructed ancestral states (see Figure \ref{fig:plate}).

\subsection{Variational Inference of the Joint Posterior}
\label{sec:variational}
We use mean-field variational inference to approximate the true joint posterior probability distribution by a new probability distribution \(q_{\phi}(\mathbf{a}_n, \mathbf{t}, \bm{\psi} | \mathbf{x}_n)\) \citep{Jordan1999,Kingma2014,Rezende2014}.
We model each latent variable by an independent approximate distribution whose parameters will be inferred using a non-linear transformation either of \(\mathbf{x}_n\), or an independent, fixed random noise \(\bm{\zeta}\). The non-linear transformations are implemented using deep neural networks (\textrm{NeuralNet}) parameterized by a set of independent and adaptable variational parameters \(\phi = \{\phi_{\mathbf{a}}, \phi_{\mathbf{t}}, \phi_{\bm{\psi}}\}\).

For each sequence \(x^m\), we infer and sample an evolutionary time variable \(\mathbf{t}^m\). We model its approximate density \(q_{\phi_{\mathbf{t}}}(\mathbf{t}^m)\) using a gamma distribution to ensure the positiveness of the samples. The parameters of the distribution (shape and rate) are produced by a non-linear transformation on uniform noise \(\bm{\zeta}_{\mathbf{t}}\) as follows:
\[
    \label{q_t}
    q_{\phi_{\mathbf{t}}}(\mathbf{t}^m) = \textrm{Gamma}(\mathbf{t}^m;\,\textrm{NeuralNet}(\bm{\zeta}_{\mathbf{t}};\, \phi_{\mathbf{t}})).
\]
Next, we infer and sample the latent variables of the Markov chain model parameters \(\bm{\psi}\) with independent approximate densities \(q_{\phi_{\bm{\psi}}}(\bm{\psi})\). 
The JC69 model does not have any free parameters to be estimated, so \(\bm{\psi}=\emptyset\). 
For the K80 model, we infer the latent variable of the transition/transversion rate ratio (\(\bm{\kappa}\)) using a gamma-based approximate distribution (\(q_{\phi_{\bm{\kappa}}}(\bm{\kappa})\)) to ensure the positiveness of the samples. Its local parameters are produced by a neural network on uniform noise \(\bm{\zeta}_{\bm{\kappa}}\) as follows:
\[
    \label{q_k}
    q_{\phi_{\bm{\kappa}}}(\bm{\kappa}) = \textrm{Gamma}(\bm{\kappa};\,\textrm{NeuralNet}(\bm{\zeta}_{\bm{\kappa}};\, \phi_{\bm{\kappa}})).
\]
In the case of the GTR model, we model the variational densities of the substitution rate parameters (\(\bm{\rho}\)) and the relative frequencies (\(\bm{\pi}\)) using Dirichlet distributions. This ensures that the sum of the sampled values is equal to one. Their concentrations are generated by a set of independent neural networks on uniform noises \(\bm{\zeta}_{\bm{\rho}}\) and \(\bm{\zeta}_{\bm{\pi}}\), respectively:
\begin{align*}
    q_{\phi_{\bm{\rho}}}(\bm{\rho}) &= \textrm{Dirichlet}(\bm{\rho}; \textrm{NeuralNet}(\bm{\zeta}_{\bm{\rho}};\, \phi_{\bm{\rho}})),\\
    q_{\phi_{\bm{\pi}}}(\bm{\pi}) &= \textrm{Dirichlet}(\bm{\pi}; \textrm{NeuralNet}(\bm{\zeta}_{\bm{\pi}};\, \phi_{\bm{\pi}})).
\end{align*}
Lastly, for each site \(x_n\), an ancestral variable \(\mathbf{a}_n\) is inferred and sampled with an approximate density \(q_{\phi_{\mathbf{a}}}(\mathbf{a}_n\,|\,\mathbf{x}_n)\)
represented by a categorical distribution over the (\(|\mathcal{A}|-1\))-simplex as follows:
\[
     \label{q_a}
    q_{\phi_{\mathbf{a}}}(\mathbf{a}_n\,|\,\mathbf{x}_n) = \textrm{Categorical}(\mathbf{a}_n; \textrm{NeuralNet}(\mathbf{x}_n;\, \phi_{\mathbf{a}})).
\]
We apply a non-linear transformation on \(\mathbf{x}_n\) to produce the local parameters of \(q_{\phi_{\mathbf{a}}}(\mathbf{a}_n\,|\,\mathbf{x}_n)\), which are a set of \(|\mathcal{A}|\) probabilities that sum to one.
Using a mean-field variational inference approach, the approximate joint posterior factorizes into:
\begin{equation}
    \label{eq:approx}
    q_{\phi}(\mathbf{a}_n, \mathbf{t}, \bm{\psi} | \mathbf{x}_n) = 
    q_{\phi_{\mathbf{a}}}(\mathbf{a}_n\,|\,\mathbf{x}_n)\,
    \prod_{m=1}^{M} q_{\phi_{\mathbf{t}}}(\mathbf{t}^m)\,
    q_{\phi_{\bm{\psi}}}(\bm{\psi}).
\end{equation}

\subsection{Generating Model Computation}
The generating model is represented by the joint density \( p(\mathbf{x}_n, \mathbf{a}_n, \mathbf{t}, \bm{\psi}) = p(\mathbf{a}_n, \mathbf{t}, \bm{\psi})\, p(\mathbf{x}_n | \mathbf{a}_n, \mathbf{t}, \bm{\psi})\), which is parameterized only by the local latent variables. We use independent prior densities for the latent variables, so \( p(\mathbf{a}_n, \mathbf{t}, \bm{\psi}) = p(\mathbf{a})\,p(\mathbf{t})\,p(\bm{\psi})\). 
To ease the computation, we apply for each prior density the same distribution type as its corresponding approximate posterior density and determine its hyper-parameters \(\varphi\).
Moreover, for each nucleotide \(x_n^m\), we use the probability transition matrix \(\mathbf{P}(\mathbf{t}^m)\) to define the likelihood function, which is the probability of evolving a character \(\mathbf{a}_n\) into \(\mathbf{x}_n^m\) during a time \(\mathbf{t}^m\), as:
\begin{equation}
\label{eq:logl}
\begin{aligned}
    \mathbf{\widehat{x}}_n^m &= \mathbf{a}_n \times \mathbf{P}(\mathbf{t}^m; \bm{\psi}),\\
    p(\mathbf{x}_n^m | \mathbf{a}_n, \mathbf{t}^m, \bm{\psi}) &= \textrm{Categorical}(\mathbf{x}_n^m ;\mathbf{\widehat{x}}_n^m).
\end{aligned}
\end{equation}
The likelihood of a site \(x_n\) is computed following a pre-order traversal. 
We call it a top-down likelihood since it includes the sampled ancestral states in its estimation.
It is different from the likelihood computed in a phylogeny, which is based on a post-order traversal \citep{felsenstein1981} and does not include sampled ancestral states.
Finally, the joint density is
\begin{equation}
    \label{eq:joint}
    p(\mathbf{x}_n, \mathbf{a}_n, \mathbf{t}, \bm{\psi}) = p(\mathbf{a})\,p(\mathbf{t})\,p(\bm{\psi})\,
    \prod_{m=1}^{M}  p(\mathbf{x}_n^m | \mathbf{a}_n, \mathbf{t}^m, \bm{\psi}).
\end{equation}

\begin{figure*}[]
    \begin{subfigure}[]{.32\textwidth}
    \caption*{100 bp}
    \includegraphics[width=\textwidth, trim={0 0.85cm 0 0}, clip]{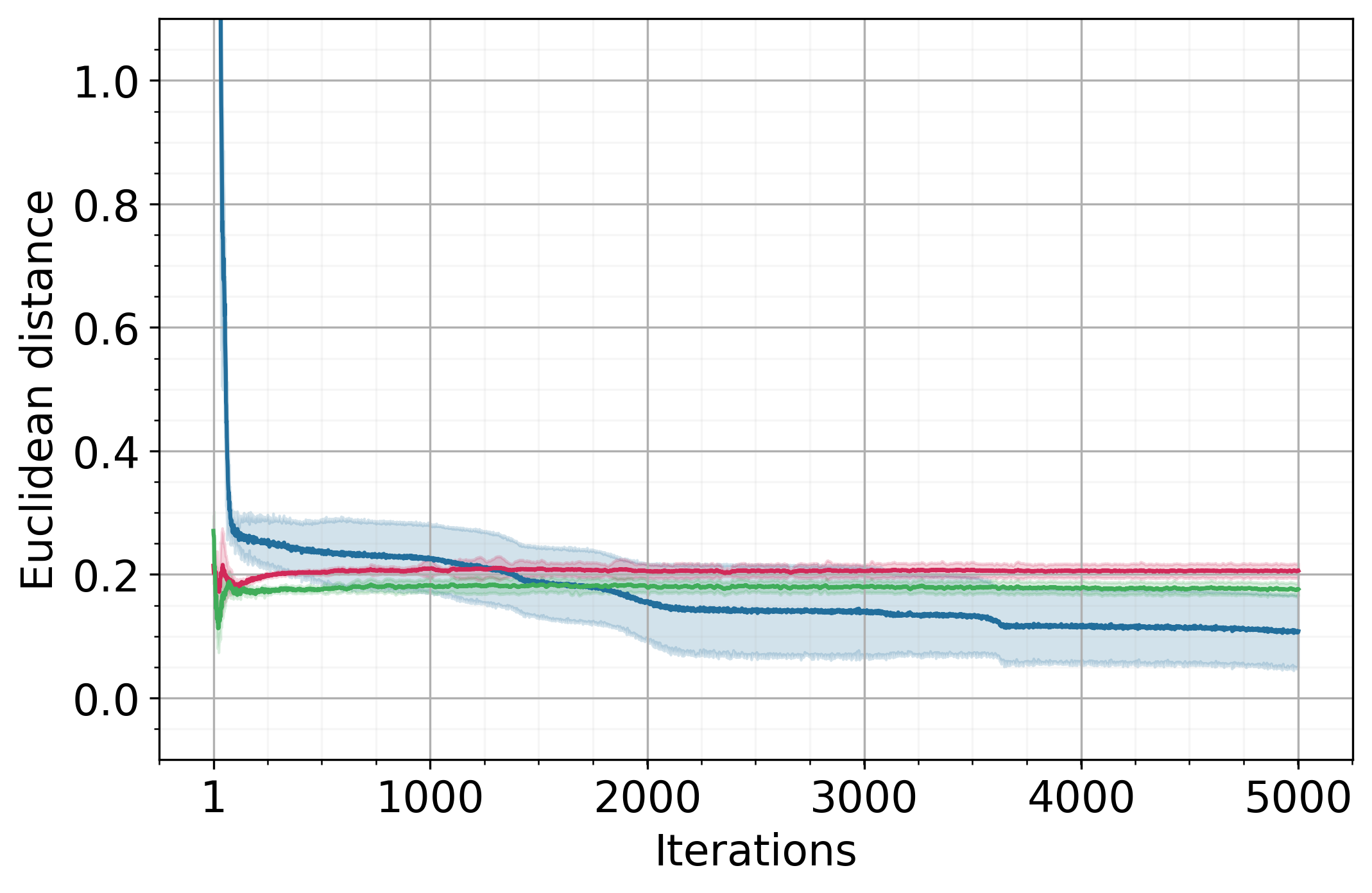}
    \includegraphics[width=\textwidth]{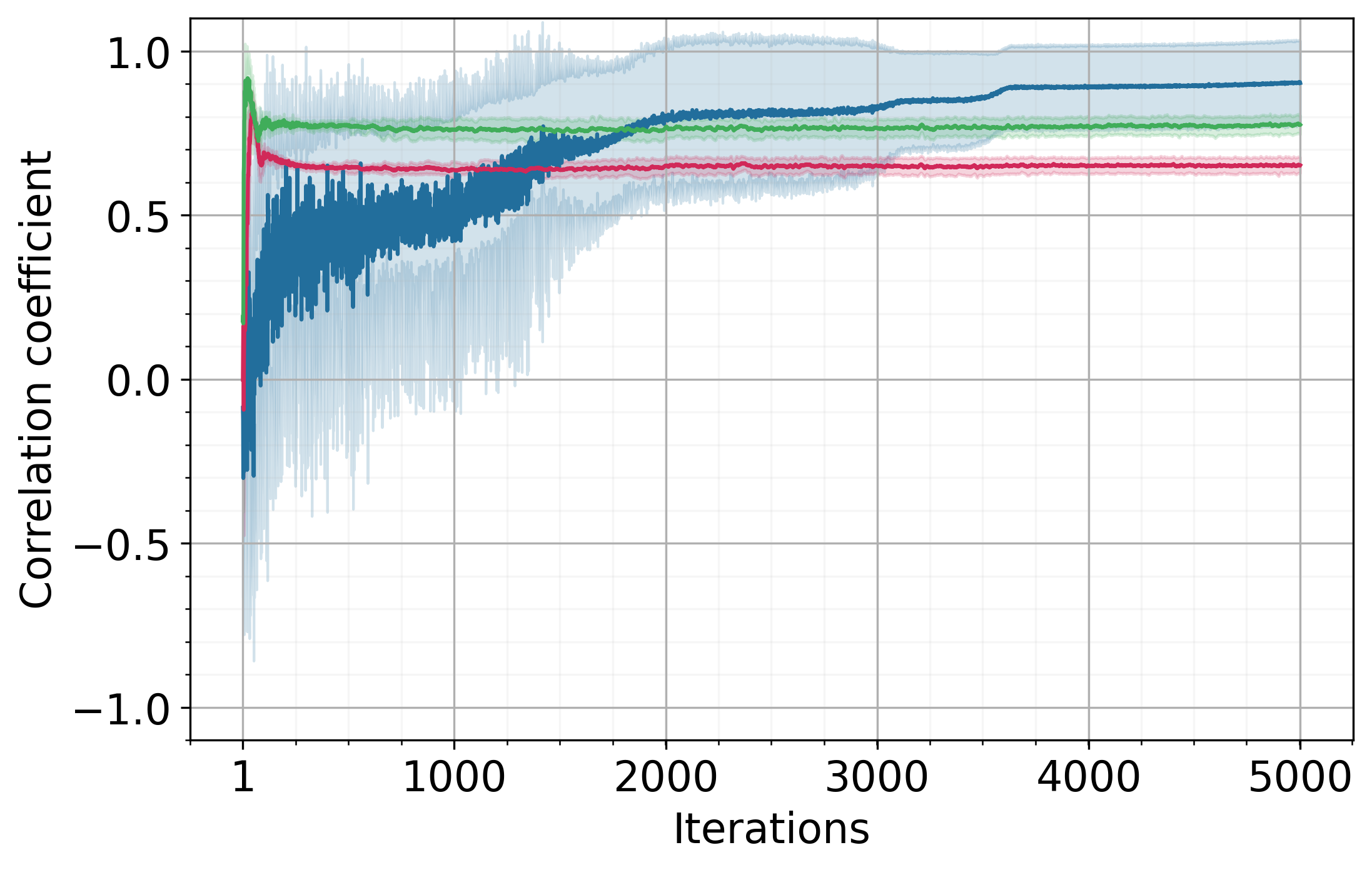}
    \end{subfigure}
    \begin{subfigure}[]{.32\textwidth}
    \caption*{1000 bp}
    \includegraphics[width=\textwidth, trim={0 0.85cm 0 0}, clip]{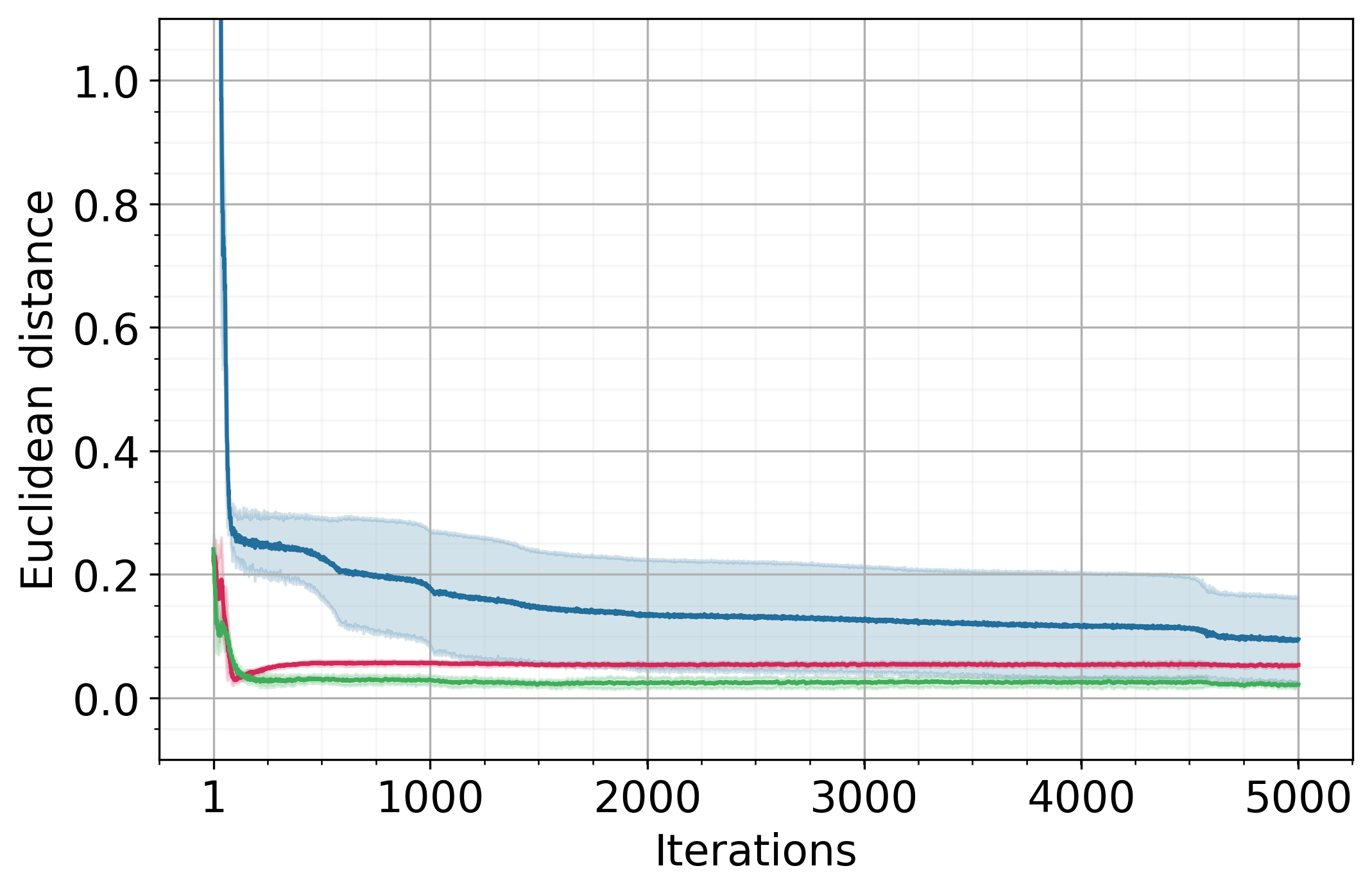}
    \includegraphics[width=\textwidth]{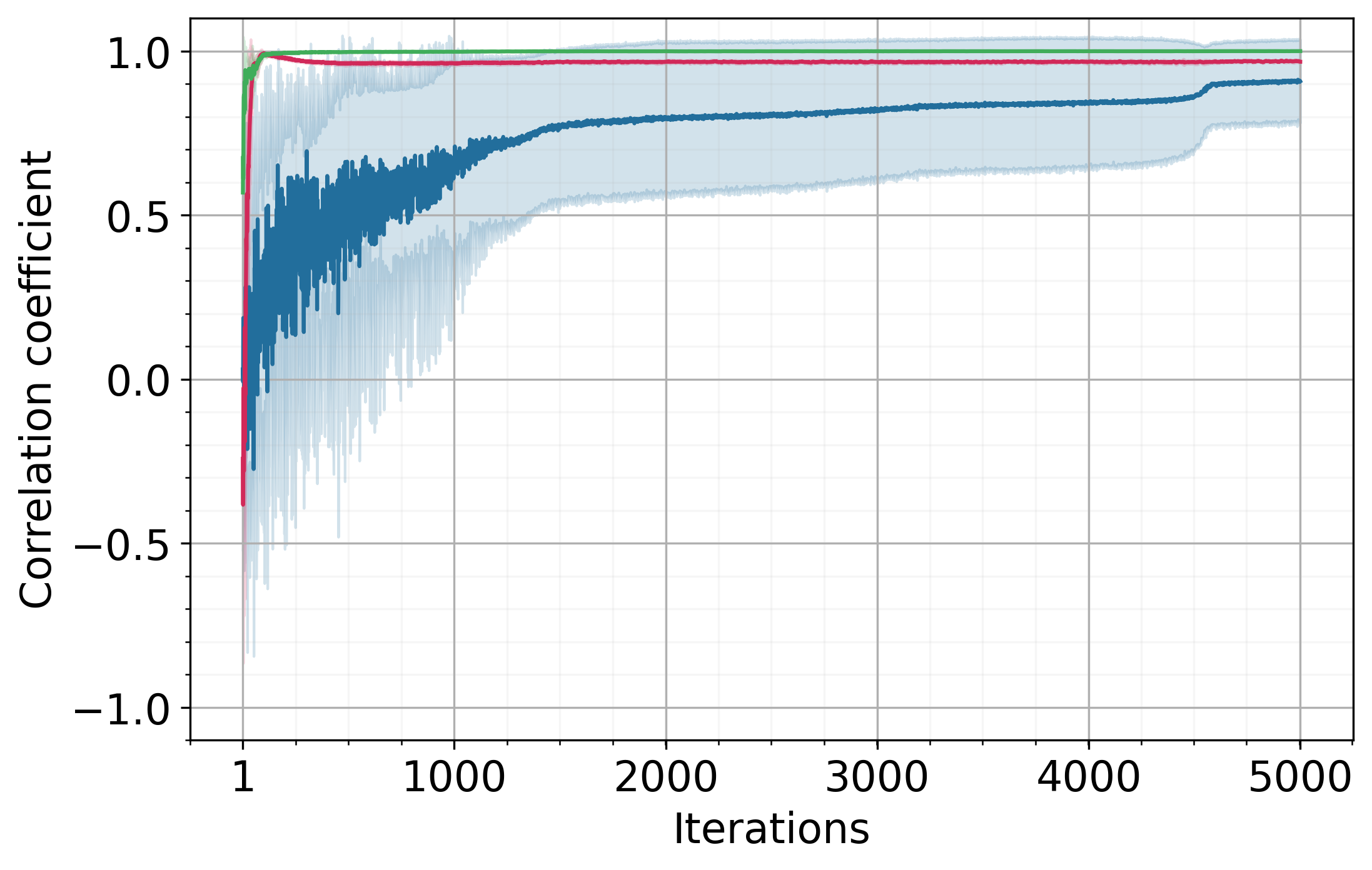}
    \end{subfigure}
    \begin{subfigure}[]{.32\textwidth}
    \caption*{5000 bp}
    \includegraphics[width=\textwidth, trim={0 0.85cm 0 0}, clip]{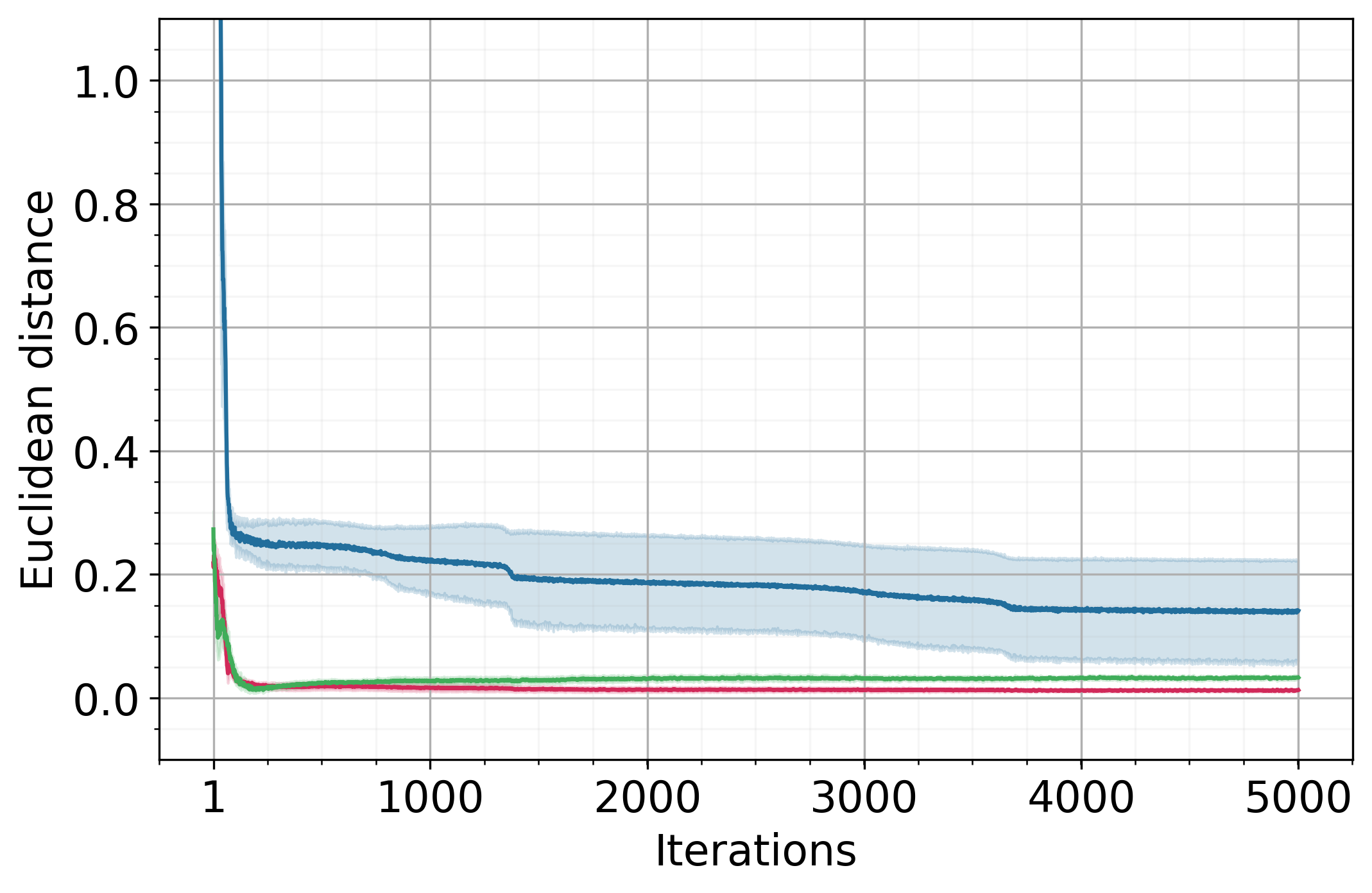}
    \includegraphics[width=\textwidth]{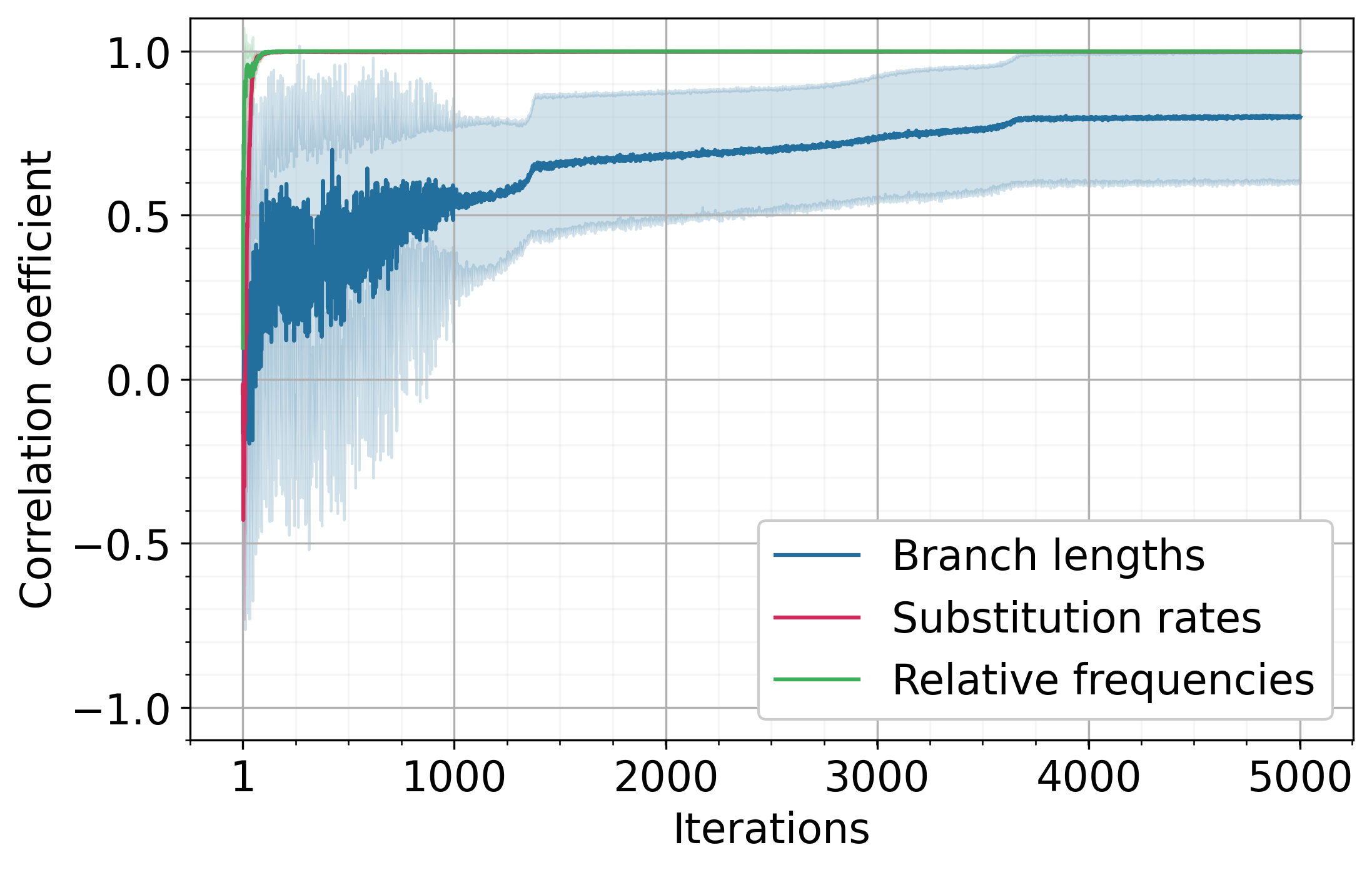}
    \end{subfigure}
    \caption{Performance of EvoVGM\_GTR model over several lengths of validation sequence alignments (100 bp, 1000 bp and 5000 bp). The GTR substitution model was used to simulate training and validation alignments of five sequences.
    The results are computed and averaged from fitting and running EvoVGM\_GTR ten times.
    They are reported in terms of Euclidean distance and Pearson correlation coefficient of estimated and actual evolutionary parameters during fitting.}
    \label{fig:gtr_nb5}
\end{figure*}

\subsection{Stochastic Estimator and Learning Algorithm}
\begin{algorithm}[b]
	\caption{Learning algorithm for \textbf{EvoVGM}}
    \label{alg:sgd}
\begin{algorithmic}
        \STATE {\bfseries Input:} Alignment \(\mathbf{X}\) of \(M\) sequences with length \(N\)
        \STATE \(\phi_{\mathbf{a}}, \phi_{\mathbf{t}}, \phi_{\bm{\psi}} \leftarrow\) initialize global variational parameters
	    \FOR{\(i \in [1 ... max\_iter]\)}
	        \STATE \(\mathbf{t}^m \leftarrow\) Sample \(M \times L\) branch latent variables (\(\phi_{\mathbf{t}}\))
	        \STATE \(\bm{\psi} \leftarrow\) Sample \(L\) evolutionary latent variables (\(\phi_{\bm{\psi}}\))
	        \STATE \(\mathbf{P}^m \leftarrow\) Compute \(M \times L\) probability transition matrices (\(\mathbf{t}^m, \bm{\psi}\))
    	    \FOR{\(n \in [1 ... N]\)}
    	        \STATE \(\mathbf{a}_n \leftarrow\) Sample \(L\) ancestor latent variable (\(\mathbf{x}_n; \phi_{\mathbf{a}}\))
    	        \STATE \(\mathbf{\hat{x}}_n \leftarrow\) Generate \(M \times L\) nucleotides (\(\mathbf{a}_n, \mathbf{P}^m\))
    	        \STATE \(\mathcal{L}_n \leftarrow\) Compute ELBO according to the equation \ref{eq:elbo} 
    	        \STATE \(\mathcal{L}  \mathrel{+}= \mathcal{L}_n\)
    	    \ENDFOR
	    \STATE \(\bm{g} \leftarrow\) Compute gradients of total ELBO (\(\mathcal{L}\))
	    \STATE \(\phi_{\mathbf{a}}, \phi_{\mathbf{t}}, \phi_{\bm{\psi}} \leftarrow\) Update parameters (\(\bm{g}\)) with gradient ascent optimizer
	    \ENDFOR
\end{algorithmic}
\end{algorithm}

Variational inference allows us to form a lower bound on the marginal likelihood of each site \(x_n\) as \(\log p(\mathbf{x}_n) \geq \mathcal{L}_n(\phi, \mathbf{x}_n)\), where \(\mathcal{L}_n\) is the evidence lower bound (ELBO) \citep{Jordan1999, Blei2017}. Putting together equations \ref{bayes}, \ref{eq:approx} and \ref{eq:joint}, we can derive the equation of the multi-sample estimator of the EvoVGM model as follows:
\begin{multline}
\label{eq:elbo}
    \mathcal{L}_n(\phi, \mathbf{x}_n) = 
    \left (\frac{1}{L} \sum_{l=1}^{L} \sum_{m=1}^{M} \log p(\mathbf{x}_n^m|\mathbf{a}_n^l, \mathbf{t}^{m,l}, \bm{\psi}^{l}) \right) \\
    - \alpha_{\texttt{KL}} \bigg( \texttt{KL}( q_{\phi_{\mathbf{a}}}(\mathbf{a}_n \,|\,\mathbf{x}_n) \parallel p(\mathbf{a})) 
    + \sum_{m=1}^{M} \texttt{KL}(q_{\phi_{\mathbf{t}}}(\mathbf{t}^m)\parallel p(\mathbf{t})) \\
    + \texttt{KL}(q_{\phi_{\bm{\psi}}}(\bm{\psi}) \parallel p(\bm{\psi})) \bigg),
\end{multline}
where \(L\) is the sampling size, \(\texttt{KL}(\cdot\parallel\cdot)\) is the Kullback–Leibler divergence, and \(\alpha_{\texttt{KL}}\) is a regularization coefficient (see the development of this equation in \ref{eq:app_elbo}). This estimator is computationally tractable because it is independent of the direct evaluation of the true joint posterior.
To maximize the ELBO and learn the global variational parameters \(\phi\), EvoVGM estimates and backpropagates the gradients for the whole data \(\mathbf{X}\) using the reparameterization trick \citep{Kingma2014} and a gradient ascent optimizer. The algorithm of EvoVGM is detailed in Algorithm \ref{alg:sgd}. It is implemented in Pytorch \citep{pytroch} and its open-source code is available at \url{https://github.com/maremita/evoVGM}.
\section{Experiments}
The evaluation of the proposed Bayesian variational model to estimate evolutionary parameters and generate sequence alignments is oriented towards assessing its consistency, effectiveness, and understanding its behavior during the training using simulated sequence alignments. Moreover, we highlight the robustness of a fine-tuned EvoVGM model using a sequence alignment of gene S of coronaviruses.
\paragraph{Sequence Alignment Simulation}
We used Pyvolve \citep{Spielman2015} to simulate the evolution of different sequence alignments with a site-wise homogeneity model and a combination of substitution models (JC69, K80 and GTR), the number of sequences (3, 4 and 5) and alignment lengths (100 bp, 1000 bp and 5000 bp). 
A site-wise homogeneity model evolves sequences from a root sequence with the same substitution model over lineages and with the same branch lengths for nucleotides. 
The sequence alignments used in the training step of the EvoVGM models were simulated with different random seeds from those used in the validation step but with the same array of evolutionary parameters.
\paragraph{Performance metrics}
We report the results of the performance of EvoVGM models in terms of the ELBO, the log likelihood (LogL), and the \texttt{KL} divergence between the approximate densities and the priors (\texttt{KL}\_qp) on the training and the validation sequence alignments.
To assess the accuracy of the estimation of the evolutionary parameters, we compute the Euclidean distance and the Pearson correlation coefficient between their arrays and those of actual parameters used in the simulation of the alignments.
\begin{figure*}[]
    \begin{subfigure}[]{.32\textwidth}
    \caption*{3 sequences}
    \includegraphics[width=\textwidth, trim={0 0.85cm 0 0}, clip]{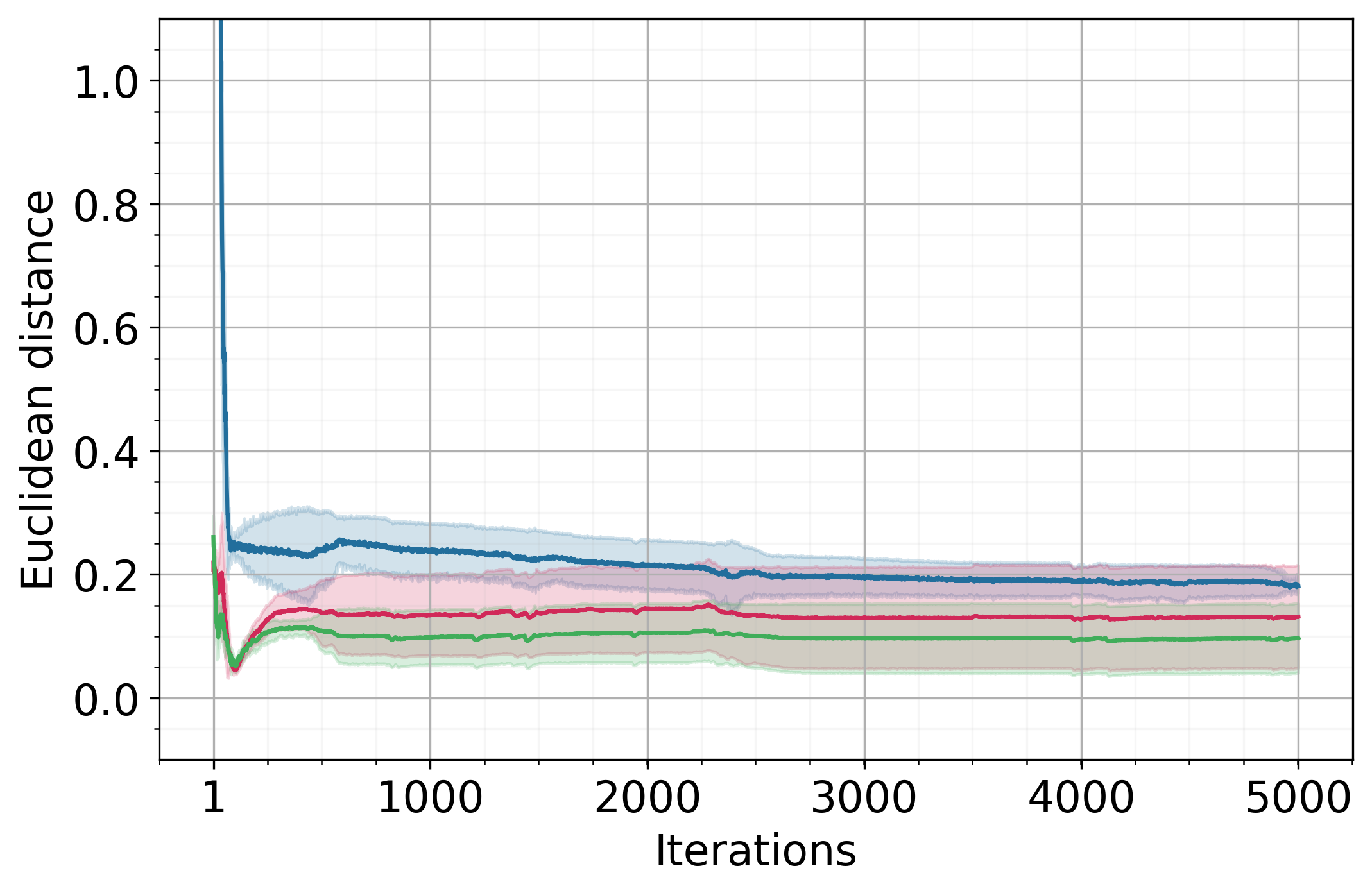}
    \includegraphics[width=\textwidth]{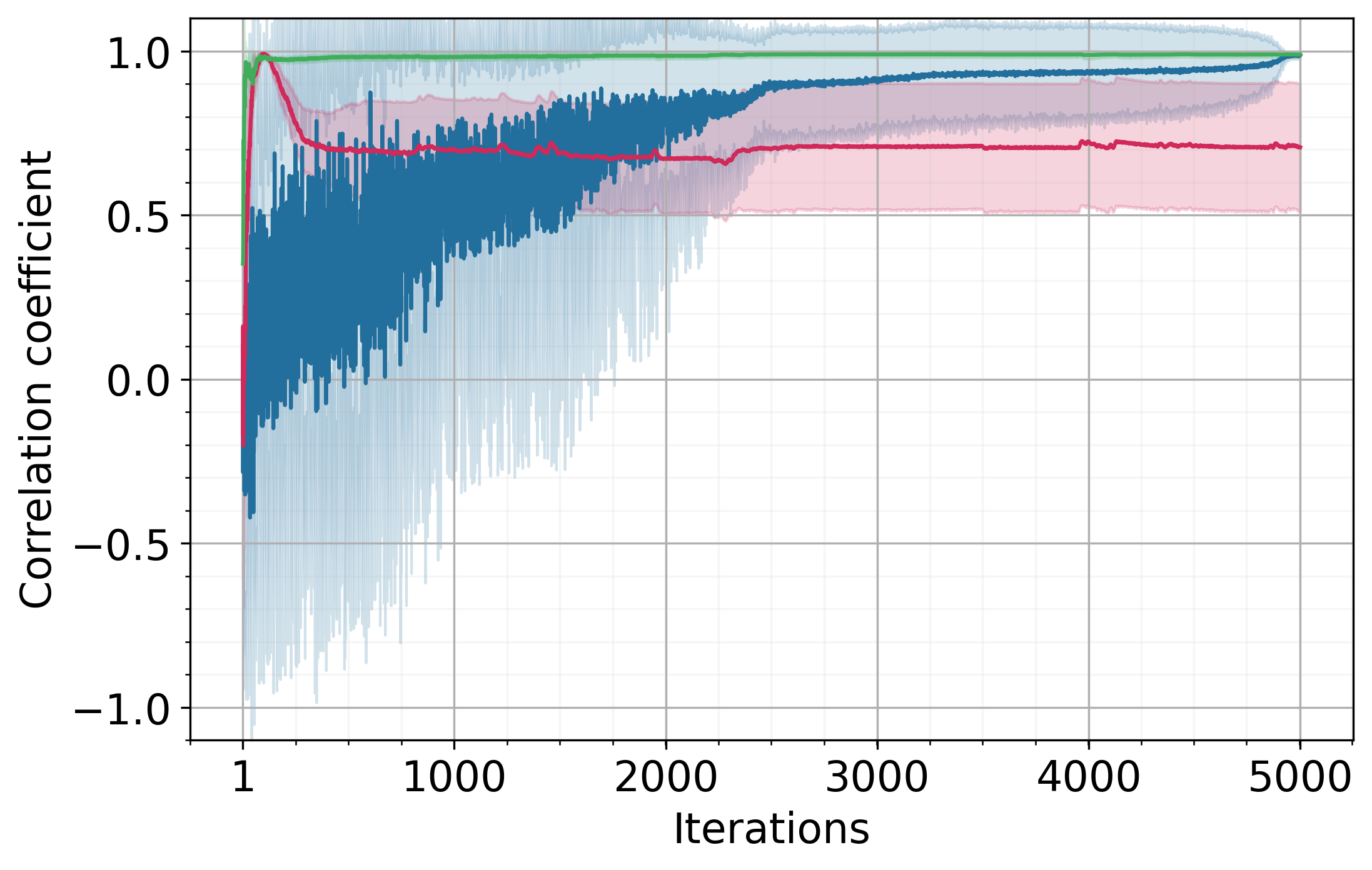}
    \end{subfigure}
    \begin{subfigure}[]{.32\textwidth}
    \caption*{4 sequences}
    \includegraphics[width=\textwidth, trim={0 0.85cm 0 0}, clip]{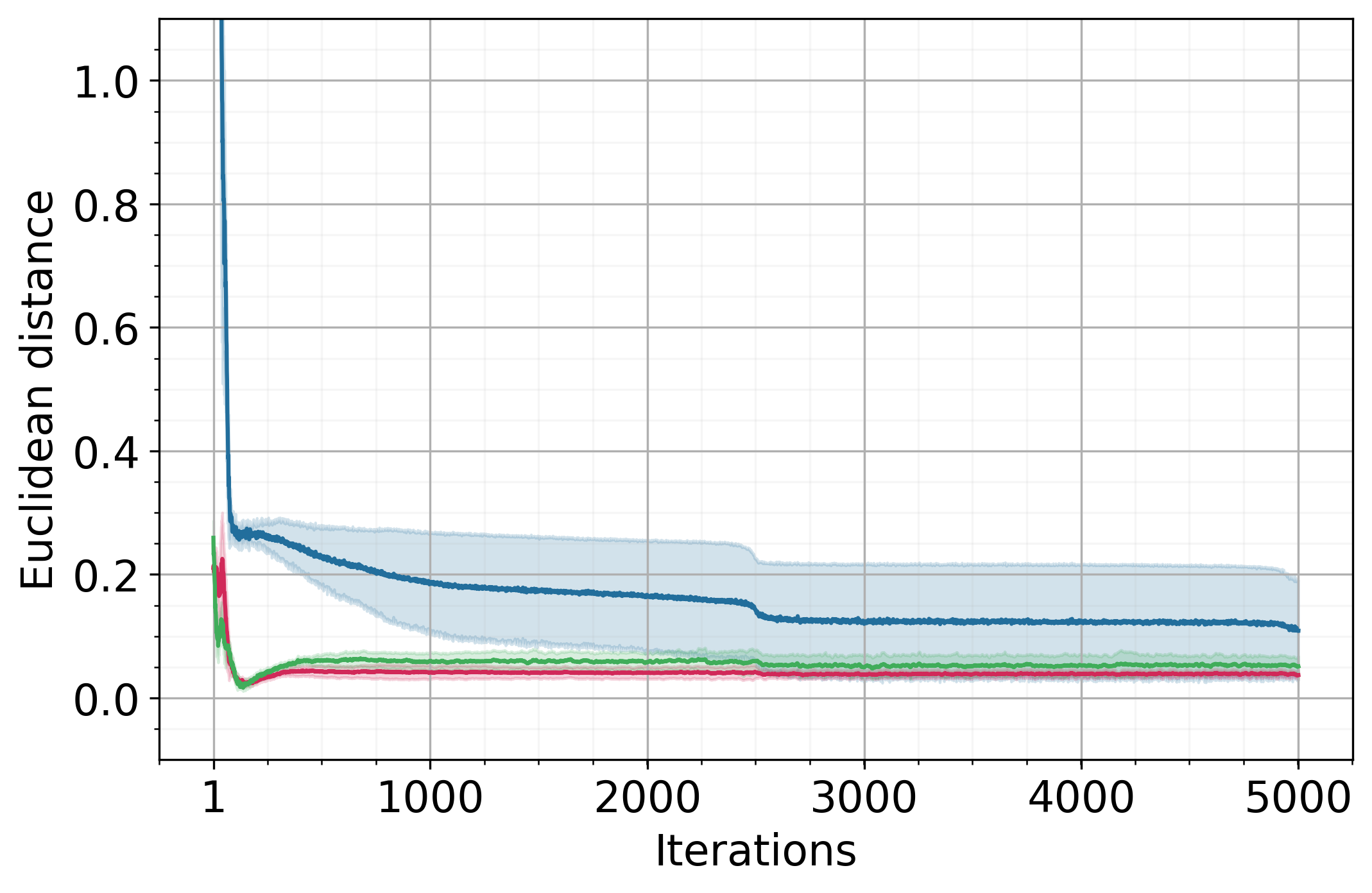}
    \includegraphics[width=\textwidth]{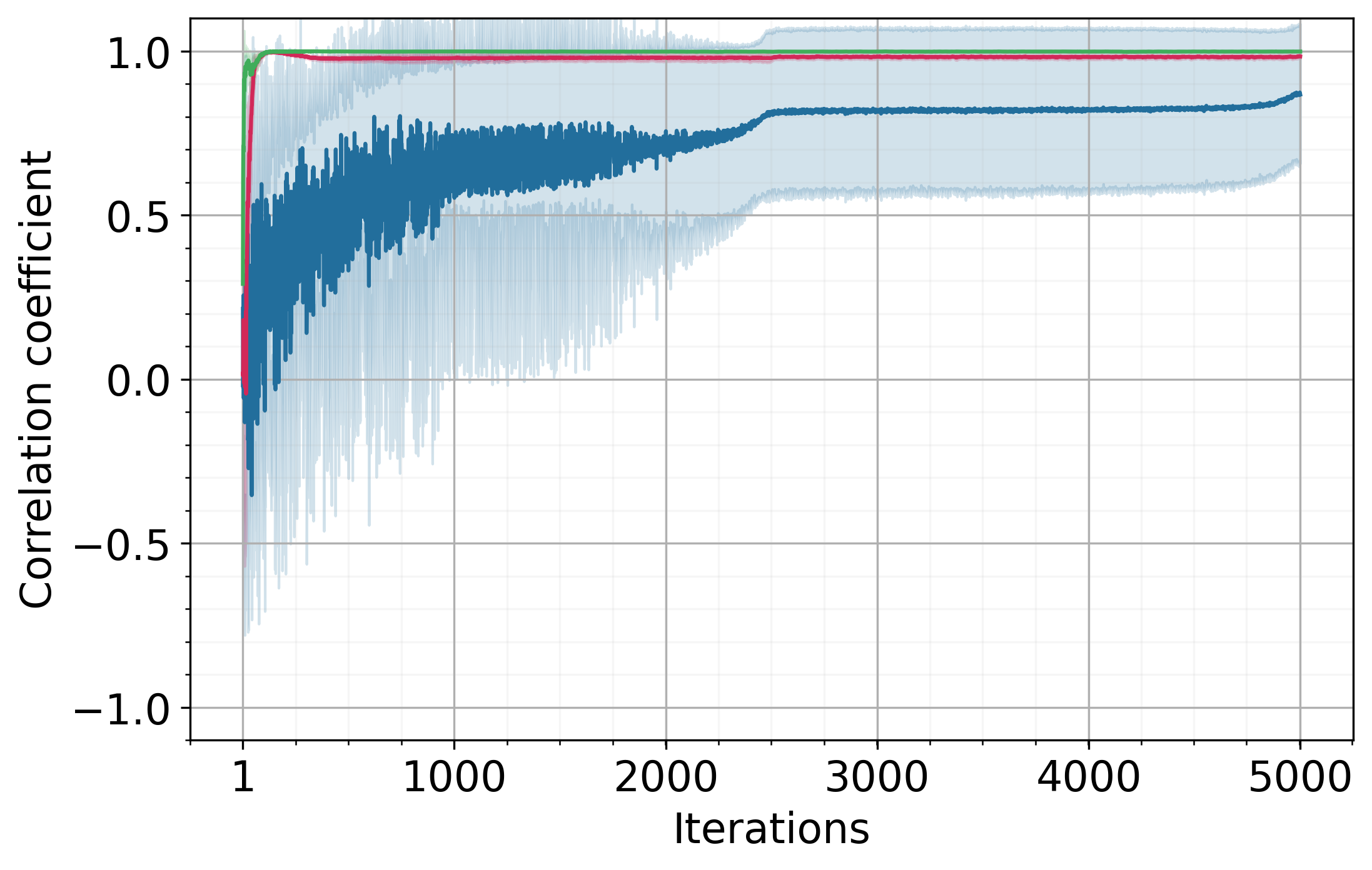}
    \end{subfigure}
    \begin{subfigure}[]{.32\textwidth}
    \caption*{5 sequences}
    \includegraphics[width=\textwidth, trim={0 0.85cm 0 0}, clip]{figures/consistency/nb5_l5000_datagtr_evogtr_val_estim_dist_exp5000.png}
    \includegraphics[width=\textwidth]{figures/consistency/nb5_l5000_datagtr_evogtr_val_estim_corr_exp5000.png}
    \end{subfigure}
    \caption{Performance of EvoVGM\_GTR model over several numbers of validation sequences (3, 4 and 5). The GTR substitution model was used to simulate training and validation alignments with a length of 5000 bp.
    The results are computed and averaged from fitting and running EvoVGM\_GTR ten times.
    They are reported in terms of Euclidean distance and Pearson correlation coefficient of estimated and actual evolutionary parameters during fitting.}
    \label{fig:gtr_l5000}
\end{figure*}

\subsection{Hyper-Parameters Fine-tuning}
First, we assessed the effects of different hyper-parameters on the convergence and the accuracy of the EvoVGM models to approximate the true distributions of the evolutionary parameters.
For each hyper-parameters setting, the model was trained ten times, using different weight initialization, on the same alignment of five 5000-bp sequences.
Based on the results of a grid search with different combinations of hyper-parameters, we defined the default components of the EvoVGM models including a set of one-hidden-layer variational encoders with a hidden size of 32. 
We set uniform hyper-parameters on the prior densities of the ancestral states, the substitution rates and the relative frequencies, and we placed independent gamma priors on the branch lengths with a prior expectation of \(0.1\).
Moreover, we used a 100-sample EvoVGM estimator with a regularization coefficient \(\alpha_{\texttt{KL}}\) equals \(10^{-3}\) and Adam optimizer \citep{Kingma2015} for stochastic gradient ascent with a learning rate of \(0.005\).
\begin{table*}[h]
\caption{Performance of trained EvoVGM\_GTR model using validation alignments simulated with different sizes. The GTR substitution model was used to simulate training and validation alignments. The results are reported in terms of Euclidean distance (DIST) and Pearson correlation coefficient (CORR and PVAL) of estimated and actual evolutionary parameters.}

\label{tab:gtr_estim}
\begin{center}
\begin{sc}
\begin{tabular}{clccccccccc}
\toprule
{} & {\(N \rightarrow\)} & \multicolumn{3}{c}{100} & \multicolumn{3}{c}{1000} & \multicolumn{3}{c}{5000} \\
{} & {\(M\)} &  Dist &  Corr &  pval &  Dist &  Corr &  pval &  Dist &  Corr &  pval \\
\midrule
\multirow{3}{*}{\shortstack{Branch\\ lengths}} & 3 & 0.300 & 0.984 & 0.114 & 0.090 & 0.980 & 0.126 & 0.097 & 0.986 & 0.107 \\
& 4 & 0.076 & 0.994 & 0.006 & 0.086 & 0.998 & 0.002 & 0.085 & 0.992 & 0.008 \\
& 5 & 0.081 & 0.986 & 0.002 & 0.069 & 0.995 & 0.000 & 0.116 & 0.962 & 0.009 \\
\midrule
\multirow{3}{*}{\shortstack{Substitution\\ rates}} & 3 & 0.621 & 0.103 & 0.846 & 0.177 & 0.668 & 0.147 & 0.129 & 0.784 & 0.065 \\
& 4 & 0.305 & 0.472 & 0.344 & 0.114 & 0.864 & 0.027 & 0.036 & 0.985 & 0.000 \\
& 5 & 0.206 & 0.652 & 0.160 & 0.053 & 0.968 & 0.002 & 0.012 & 0.998 & 0.000 \\
\midrule
\multirow{3}{*}{\shortstack{Relative\\ frequencies}} & 3 & 0.190 & 0.941 & 0.059 & 0.084 & 0.991 & 0.009 & 0.095 & 0.992 & 0.008 \\
& 4 & 0.125 & 0.891 & 0.109 & 0.090 & 0.996 & 0.004 & 0.050 & 0.999 & 0.001 \\
& 5 & 0.176 & 0.775 & 0.225 & 0.022 & 1.000 & 0.000 & 0.033 & 0.999 & 0.001 \\
\bottomrule
\end{tabular}
\end{sc}
\end{center}
\end{table*}

Figures \ref{fig:gtr_kl}, \ref{fig:gtr_hs}, \ref{fig:gtr_ns} and \ref{fig:gtr_lr}, in the Supplemental Results section, highlight the convergence and the performance of EvoVGM\_GTR model (implementing the GTR substitution model) trained with multiple values of the coefficient \(\alpha_{\texttt{KL}}\), the size of the hidden layers, the sampling size, and the learning rate, respectively. 
In general, EvoVGM\_GTR models converge faster when the \(\alpha_{\texttt{KL}}\) coefficient is lower, and the number of hidden layers and the learning rate are larger.
The sample size does not affect the overall convergence. However, a small sample size induces a substantial variance in the estimator.

\begin{table*}[h]
\caption{Log likelihood estimates of EvoVGM models using validation alignments of five sequences with a length of 5000 bp. JC69, K80 and GTR substitution models were used to simulate training and validation alignments. The estimates are computed and averaged from fitting and running the models ten times.}

\label{tab:logl_comp}
\begin{center}
\begin{sc}
\begin{tabular}{lcccccc}
\toprule
{} & \multicolumn{2}{c}{jc69} & \multicolumn{2}{c}{k80} & \multicolumn{2}{c}{gtr} \\
{} &       Mean &     STD &       Mean &     STD &       Mean &     STD \\
\midrule
Actual & -17249.830 &         & -17024.340 &         & -15818.739 &         \\
EvoVGM\_JC69 & -17209.913 & 142.128 & -17287.278 & 185.441 & -16491.810 & 125.664 \\
EvoVGM\_K80  & -17203.100 & 151.758 & -17007.724 & 133.294 & -16495.530 & 175.024 \\
EvoVGM\_GTR  & -17204.540 & 121.459 & -17014.296 & 151.457 & -15540.730 & 126.673 \\
\bottomrule
\end{tabular}
\end{sc}
\end{center}
\end{table*}

\subsection{Assessing Consistency on Simulated Data}
Next, we analyzed the consistency and the effectiveness of the EvoVGM\_GTR model on sequence alignments simulated with different sizes in terms of length and number of sequences.
We built the model using the same default configuration and the same hyper-parameters defined in the previous analysis.
Figure \ref{fig:gtr_nb5} shows the accuracy of EvoVGM\_GTR during its training in terms of the Euclidean distance and the correlation coefficient of the estimated evolutionary parameters using validation sequence alignments of five sequences with lengths of 100 bp, 1000 bp and 5000 bp.
Conversely, Figure \ref{fig:gtr_l5000} shows the accuracy of the model in approximating the parameters on validation sequence alignments of three, four and five sequences with a length of 5000 bp.
Usually, the parameter approximation is improved when the number of sequences is higher, and the alignments are longer.
On the one hand, branch lengths suffer from high variance estimation and slow evolution at the beginning of the training across all datasets. However, its variance and accuracy improve with more training iterations. 
On the other hand, the estimation of the substitution rates and the relative frequencies converges faster and has low variance. It is more accurate when the model is trained with larger sequence alignments.

We present in Table \ref{tab:gtr_estim} the performance of trained EvoVGM\_GTR models in approximating the evolutionary parameters from new multiple alignments, which were simulated with the same set of evolutionary parameters used in the training step. 
We noted that the estimated branch lengths are close and strongly correlated to their actual values for all datasets except the smallest.
Also, we noted that EvoVGM\_GTR estimates better the relative frequencies than the substitution rates in the small datasets. However, as the datasets get larger, the approximation of the substitution rates gets better.
\begin{figure}[h]
    \begin{subfigure}[]{.235\textwidth}
    \caption*{\small EvoVGM\_JC69}
    \includegraphics[width=\textwidth, trim={0 0.85cm 0 0}, clip]{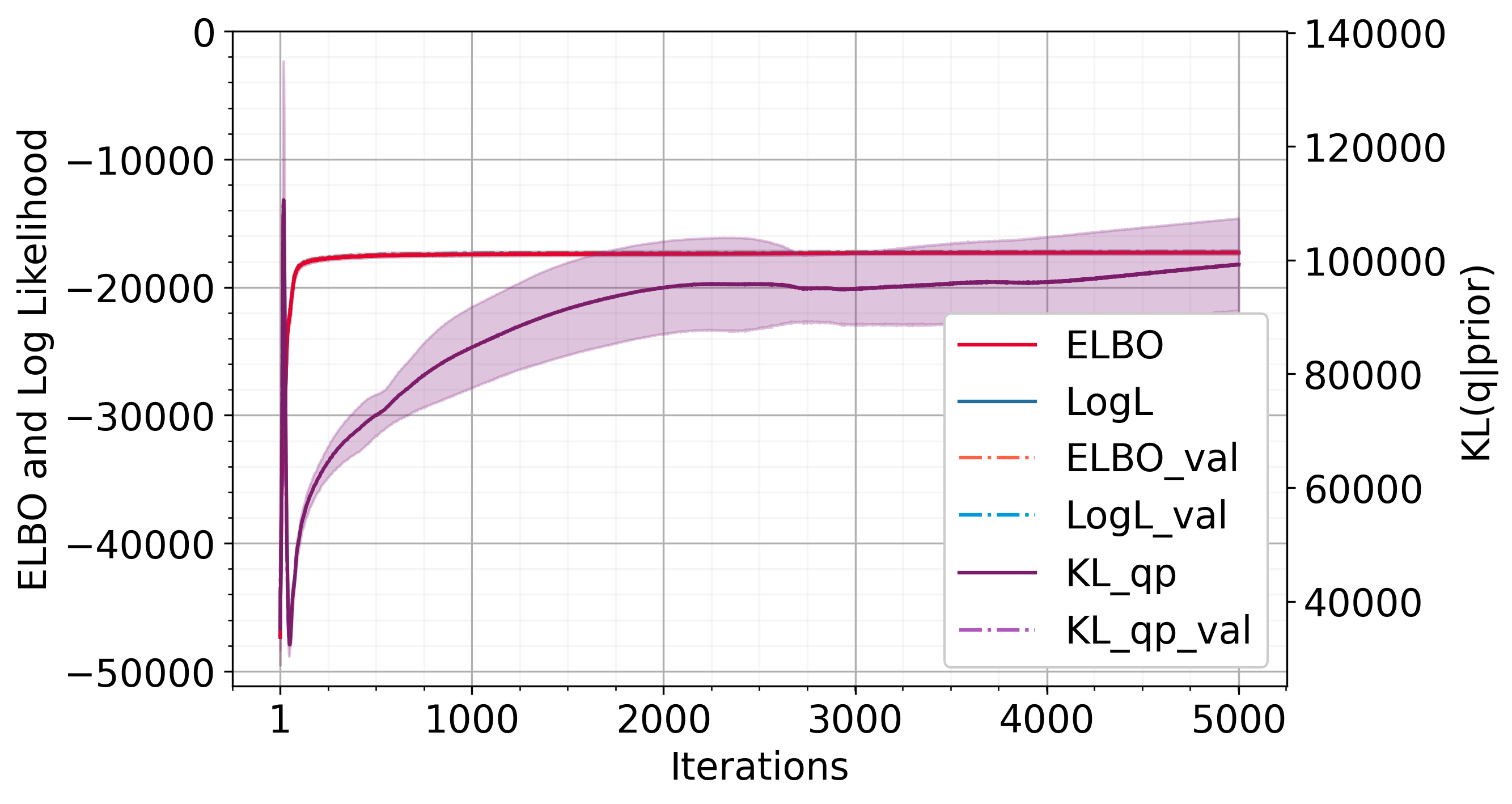}
    \includegraphics[width=\textwidth, trim={0 0.85cm 0 0}, clip]{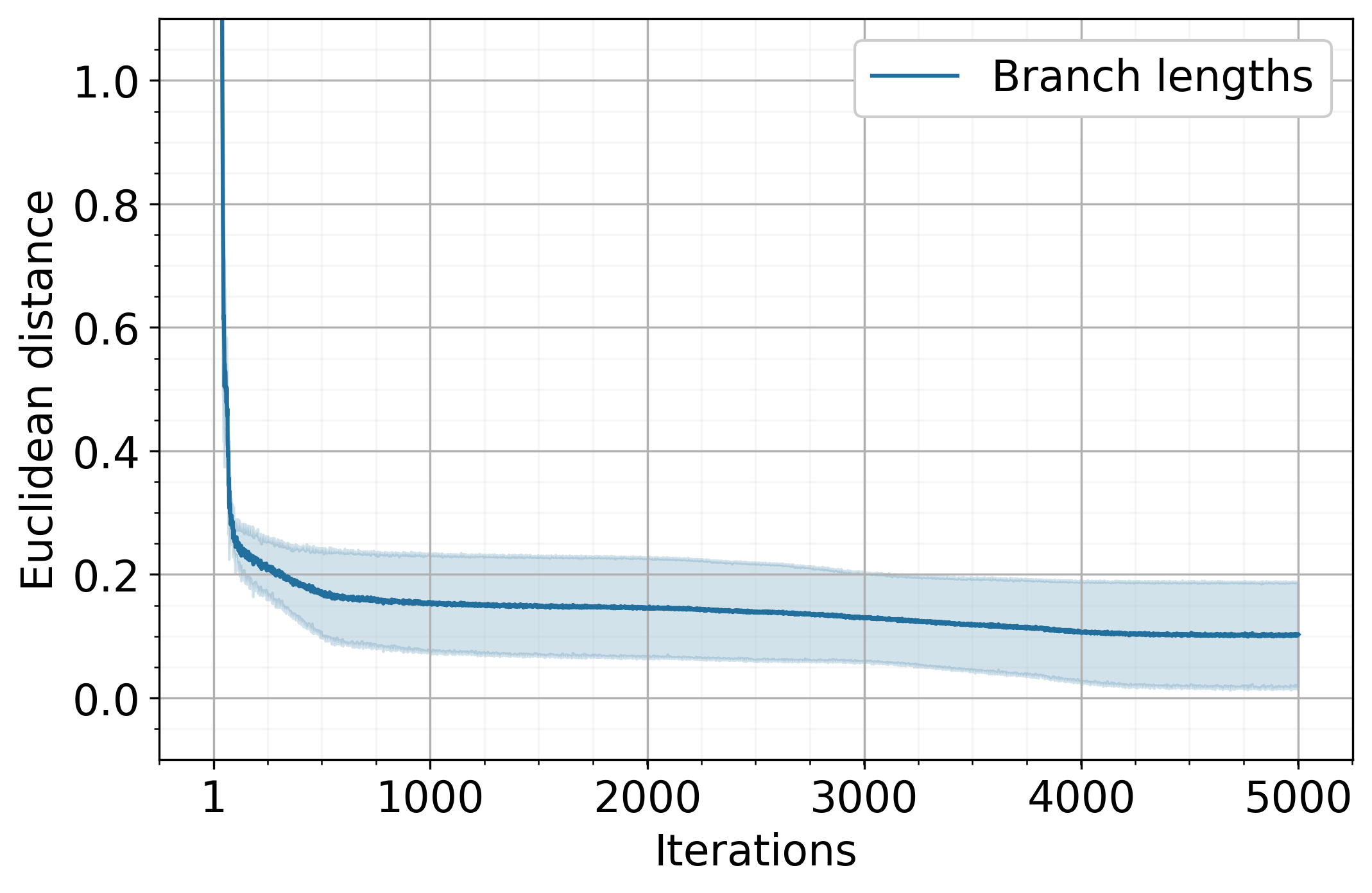}
    \includegraphics[width=\textwidth]{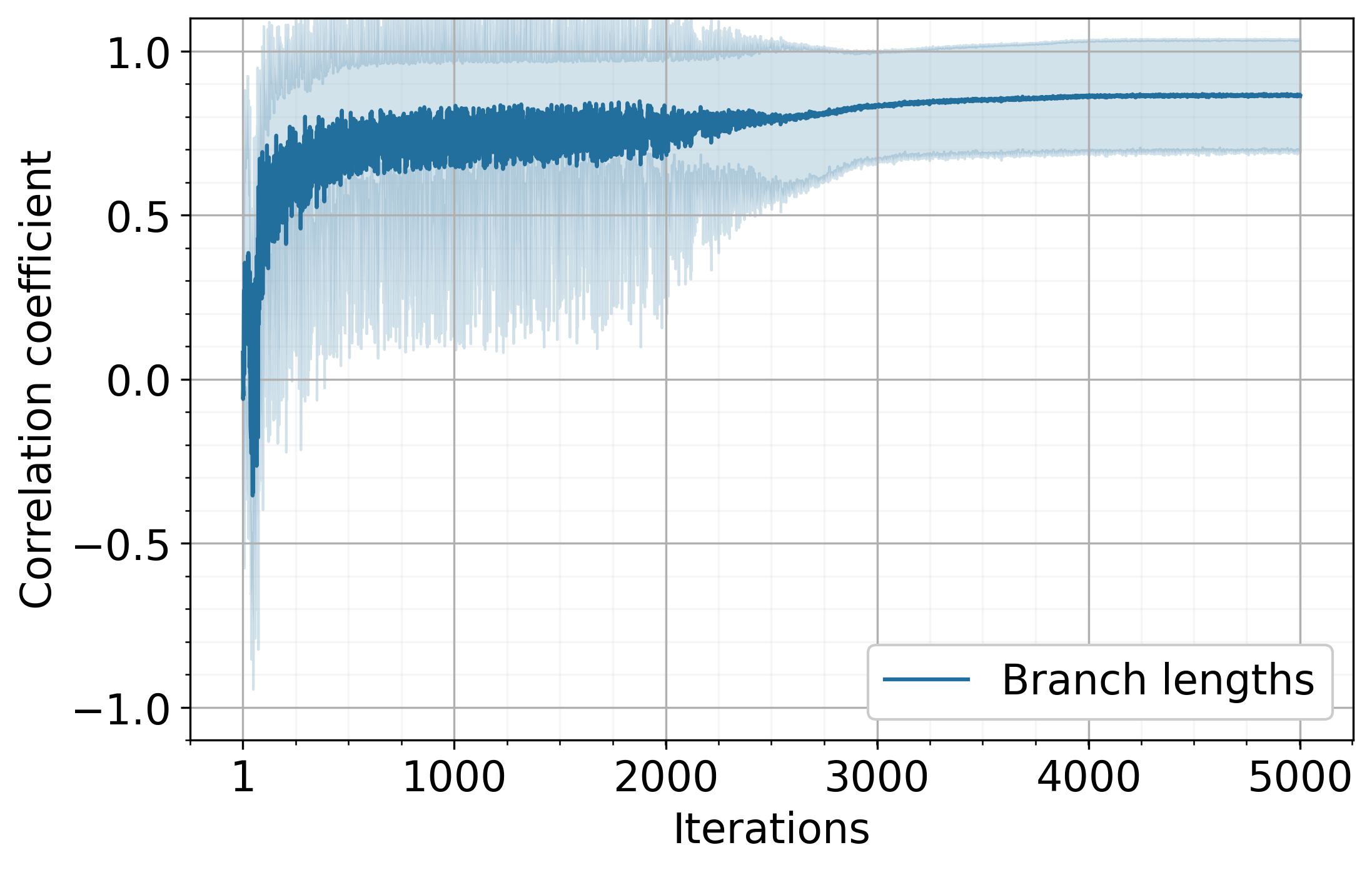}
    \end{subfigure}
    \begin{subfigure}[]{.235\textwidth}
    \caption*{\small EvoVGM\_K80}
    \includegraphics[width=\textwidth, trim={0 0.85cm 0 0}, clip]{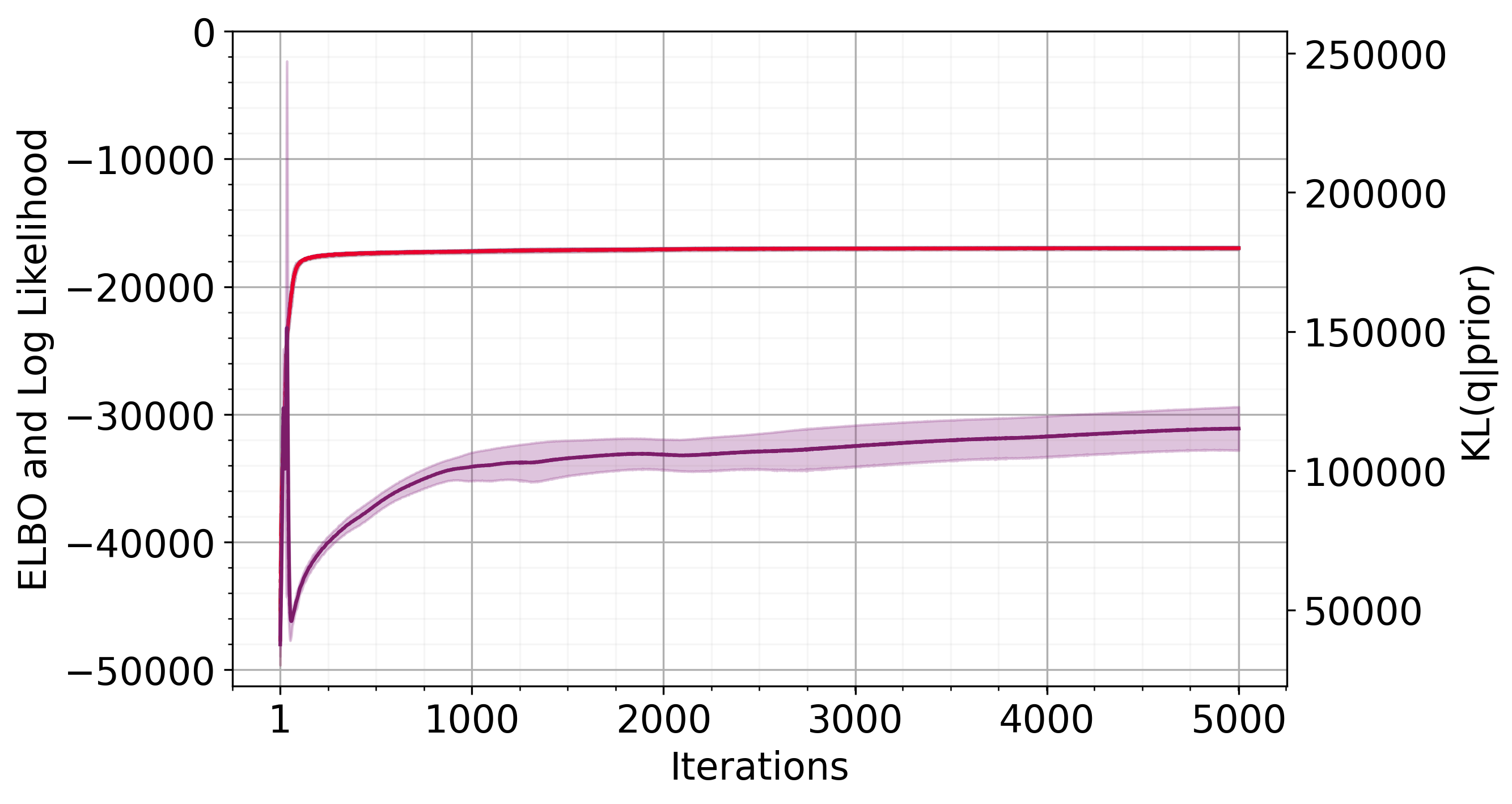}
    \includegraphics[width=\textwidth, trim={0 0.85cm 0 0}, clip]{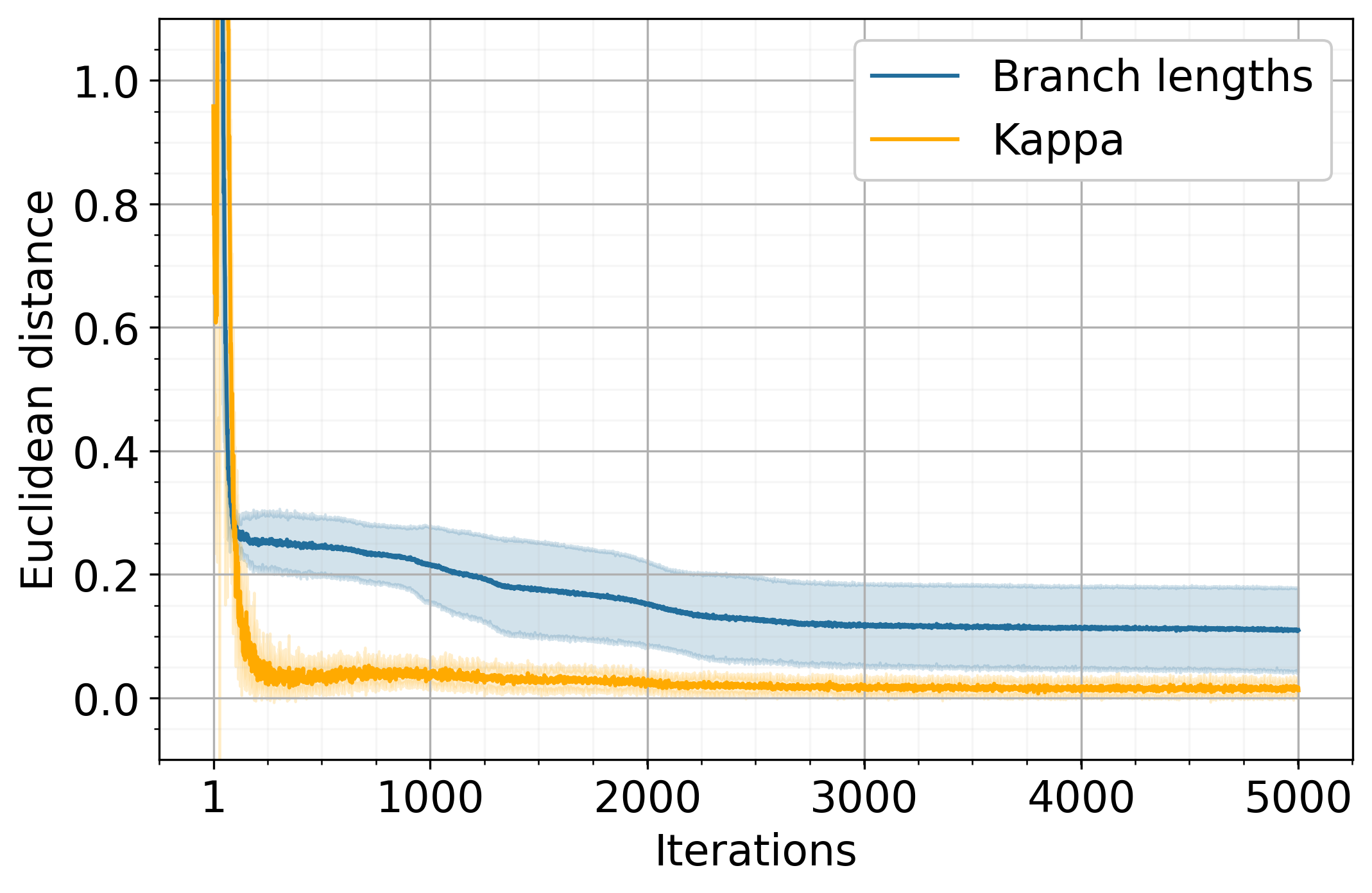}
    \includegraphics[width=\textwidth]{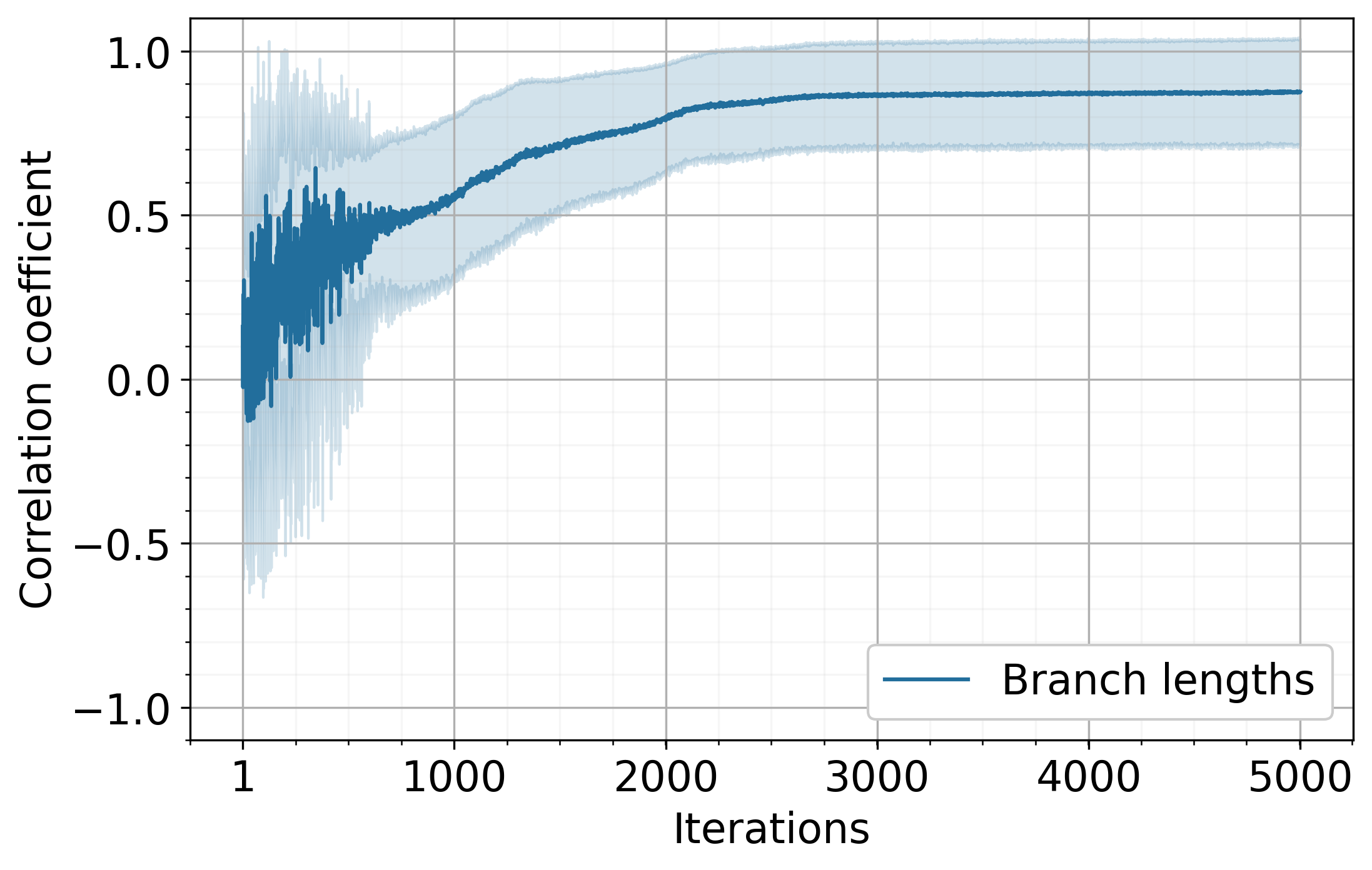}
    \end{subfigure}
    \caption{Convergence and performance of EvoVGM\_JC69 and EvoVGM\_K80 models using alignments of five sequences with a length of 5000 bp. JC69 and K80 substitution models were used to simulate training and validation alignments. The results are computed and averaged from fitting and running the models ten times.}
    \label{fig:jc69_k80}
\end{figure}

Additionally, we evaluated the convergence and the accuracy of EvoVGM\_JC69 and EvoVGM\_K80, two variants of the EvoVGM model implementing JC69 and K80 substitution models, respectively.
Each model was fitted ten times with different weight initialization on the exact sequence alignment.
Figure \ref{fig:jc69_k80} and Table \ref{tab:logl_comp} show the results of the models trained and evaluated with alignments of five sequences with a length of 5000 bp.
All models converge to values closer to or higher than the actual log likelihood of the data, which is calculated with equation \ref{eq:logl}.
Moreover, they were able to approximate the branch lengths even when trained with datasets simulated with a different substitution model (Tables \ref{tab:jc_b} and \ref{tab:k80_b}).

\subsection{Estimating Evolutionary Parameters on Real Alignment}
Finally, we analyzed a real dataset to assess the robustness of the estimation provided by the variational generative model EvoVGM. The dataset was recovered from \cite{Samson2022}, and it consists of six sequences of Gene S of coronaviruses. We used the NGPhylogeny.fr platform \citep{Lemoine2019} to build a multiple sequence alignment with MAFFT 7.407 \citep{Katoh2013}. The alignment has a length of 3688 bp after cleaning the sequences from gaps using Gblocks 0.91.1 \citep{Talavera2007}.\\
We applied EvoVGM\_GTR model to the dataset using the same configuration of the variational encoders and the hyper-parameters defined previously. We set gamma priors on branch lengths with a prior expectation of \(0.01\). We found that using \(\alpha_{\texttt{KL}}\) with a value of \(0.1\) gives better estimations but with higher variance. We trained EvoVGM\_GTR over 5000 iterations and replicated it ten times.
Furthermore, we compared the estimations of EvoVGM\_GTR model with those of MrBayes 3.2.7, a Bayesian phylogenetic inference program \citep{Huelsenbeck2001mrbayes}. We ran MrBayes with four chains and two runs for one million iterations and sampling every 500 iterations.

\begin{figure}[]
    \begin{subfigure}[]{0.235\textwidth}
    \centering
    \includegraphics[width=\textwidth]{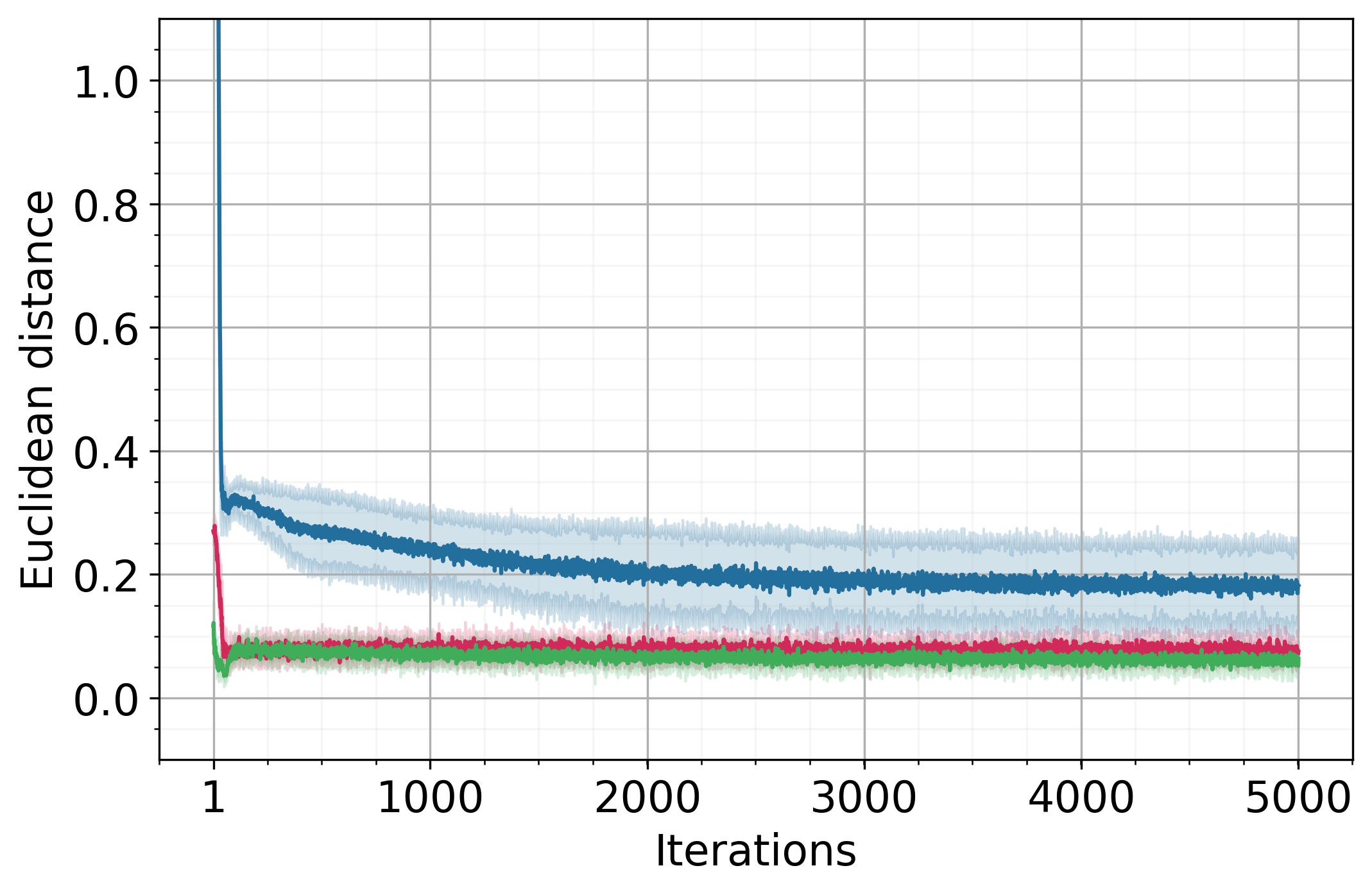}
    \end{subfigure}
    \begin{subfigure}[]{0.235\textwidth}
    \centering
    \includegraphics[width=\textwidth]{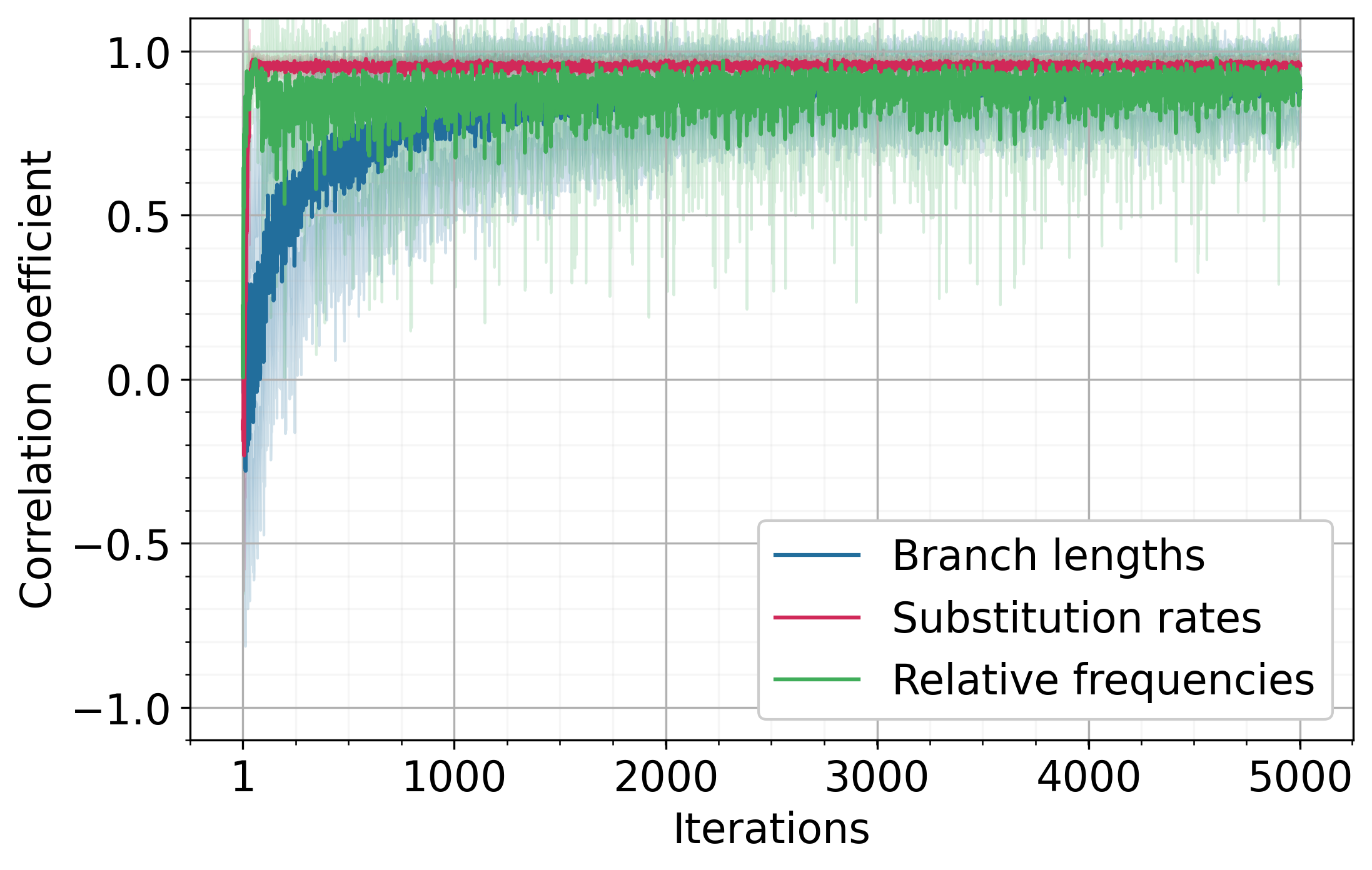}
    \end{subfigure}
    \caption{Comparison of EvoVGM\_GTR with MrBayes. The results are averaged from fitting EvoVGM\_GTR ten times. They are reported in terms of Euclidean distance and correlation coefficient of EvoVGM\_GTR and MrBayes estimated evolutionary parameters.}
    \label{fig:cov}
\end{figure}

In Figure \ref{fig:cov}, we show the evolution of the Euclidean distance and the correlation coefficient between the estimations of EvoVGM\_GTR and those of MrBayes during the training of EvoVGM\_GTR. The estimations of the branch lengths differ from those of MrBayes with a distance lower than 0.2 but with high variance. Furthermore, EvoVGM\_GTR managed to estimate the substitution rates and the relative frequencies with low-variance values closer to the estimations of MrBayes.




\section{Conclusion}
In this work, we show that a deep variational Bayesian generative method could constitute a feasible option to approximate the true parameters of an evolutionary model and generate the associated sequence alignment. 
The implementation of this method, EvoVGM, estimates the branch lengths, the ancestral states, and the substitution model parameters from a multiple sequence alignment.
We assessed its consistency and effectiveness using sequence alignments simulated with different sizes. In general, the EvoVGM model needs a few thousand iterations to converge. 
It tends to be accurate with low variance in estimating the evolutionary parameters using fine-tuned hyper-parameters.
Moreover, it provides an effective way of estimating the parameters for different substitution models such as JC69, K80, and GTR. The generalization to other models like HKY is also straightforward.  
For future work, many extensions could be explored to improve the EvoVGM model, such as considering a prior tree topology, investigating the influence of the priors on inference, and allowing parameter heterogeneity across sites and lineages.

\begin{acks}
We would like to thank Golrokh Vitae, Hayda Almeida, Maia Kaplan, Dylan Lebatteux, Mathieu Blanchette and Vladimir Makarenkov for their helpful discussions.\\
This research was enabled in part by support provided by Calcul Qu\'ebec and the Digital Research Alliance of Canada. 
It has also been supported by the Natural Sciences and Engineering Research Council of Canada (NSERC), the Fonds de recherche du Qu\'ebec - Nature et technologies (FRQNT), G\'enome Qu\'ebec and Genome Canada for the grants to ABD. AMR received NSERC and FRQNT scholarships during the development of this work.
\end{acks}

\bibliographystyle{ACM-Reference-Format}
\bibliography{2022_evovgm}
\appendix
\renewcommand\thefigure{\thesection.\arabic{figure}}
\renewcommand\thetable{\thesection.\arabic{table}}
\setcounter{equation}{0}
\setcounter{figure}{0}
\setcounter{table}{0}

\section{Appendix}
\subsection{Development of the ELBO \(\mathcal{L}(\phi, \mathbf{X})\)}
\label{eq:app_elbo}

\begin{equation*}
\begin{aligned}
\log p(\mathbf{X}) &= \sum_{n=1}^{N} \log p(\mathbf{x}_n)\\
             	&= \sum_{n=1}^{N} \mathbb{E}_{q_{\phi}(\mathbf{a}_n, \mathbf{t}, \bm{\psi} |\,\mathbf{x}_n)} \left [ \log \frac{p(\mathbf{x}_n, \mathbf{a}_n, \mathbf{t}, \bm{\psi})}{p(\mathbf{a}_n, \mathbf{t}, \bm{\psi}\,|\,\mathbf{x}_n)} \right ]\\
             	&= \sum_{n=1}^{N} \mathbb{E}_{q_{\phi}(\mathbf{a}_n, \mathbf{t}, \bm{\psi}\,|\,\mathbf{x}_n)} \left[ \log \frac{p(\mathbf{x}_n, \mathbf{a}_n, \mathbf{t}, \bm{\psi})}{q_{\phi}(\mathbf{a}_n, \mathbf{t}, \bm{\psi}\,|\,\mathbf{x}_n)} \, \frac{q_{\phi}(\mathbf{a}_n, \mathbf{t}, \bm{\psi}\,|\,\mathbf{x}_n)}{p(\mathbf{a}_n, \mathbf{t}, \bm{\psi}\,|\,\mathbf{x}_n)} \right ]\\
             	&= \sum_{n=1}^{N} \mathbb{E}_{q_{\phi}} \left [ \log \frac{p(\mathbf{x}_n, \mathbf{a}_n, \mathbf{t}, \bm{\psi})}{q_{\phi}(\mathbf{a}_n, \mathbf{t}, \bm{\psi}\,|\,\mathbf{x}_n)} \right ] + \mathbb{E}_{q_{\phi}} \left [ \log \frac{q_{\phi}(\mathbf{a}_n, \mathbf{t}, \bm{\psi}\,|\,\mathbf{x}_n)}{p(\mathbf{a}_n, \mathbf{t}, \bm{\psi}\,|\,\mathbf{x}_n)}  \right ]\\
             	&= \underbrace{\sum_{n=1}^{N} \mathcal{L}_n(\phi, \mathbf{x}_n)} + \sum_{n=1}^{N} \texttt{KL}\left( q_{\phi}(\mathbf{a}_n, \mathbf{t}, \bm{\psi}\,|\,\mathbf{x}_n) \parallel p(\mathbf{a}_n, \mathbf{t}_n, \bm{\psi}\,|\,\mathbf{x}_n) \right)\\
                &\geq \hphantom{\sum_{n=1}^{N}}\mathcal{L}(\phi, \mathbf{X}).\\ 
\mathcal{L}(\phi, \mathbf{X}) &= \sum_{n=1}^{N} \mathcal{L}_n(\phi, \mathbf{x}_n)\\
             	&= \sum_{n=1}^{N} \mathbb{E}_{q_{\phi}} \left [ \log \frac{p(\mathbf{x}_n, \mathbf{a}_n, \mathbf{t}, \bm{\psi})}{q_{\phi}(\mathbf{a}_n, \mathbf{t}, \bm{\psi}\,|\,\mathbf{x}_n)} \right ] \\
             	&= \sum_{n=1}^{N} \mathbb{E}_{q_{\phi}} \bigg [\log p(\mathbf{x}_n | \mathbf{a}_n, \mathbf{t}, \bm{\psi}) + \log p(\mathbf{a}) + \log p(\mathbf{t}) + \log p(\bm{\psi}) \\
             	&\hphantom{\sum_{n=1}^{N}}- \log q_{\phi_{\mathbf{a}}}(\mathbf{a}_n \,|\,\mathbf{x}_n) - \log q_{\phi_{\mathbf{t}}}(\mathbf{t}) - \log q_{\phi_{\bm{\psi}}}(\bm{\psi}) \bigg ]\\
              	&= - N \left( \mathbb{E}_{q_{\phi}}\left[ \log p(\bm{\psi}) - \log q_{\phi_{\bm{\psi}}}(\bm{\psi}) \right ] 
              	+ \mathbb{E}_{q_{\phi}}\left[ \log p(\mathbf{t}) - \log q_{\phi_{\mathbf{t}}}(\mathbf{t}) \right ] \right)\\
              	&\hphantom{= \;}+ \sum_{n=1}^{N} 
                \mathbb{E}_{q_{\phi}}\left[\log p(\mathbf{x}_n|\mathbf{a}_n,\mathbf{t},\bm{\psi})\right ]
                + \mathbb{E}_{q_{\phi}}\left[\log p(\mathbf{a}) - \log q_{\phi_{\mathbf{a}}}(\mathbf{a}_n \,|\,\mathbf{x}_n) \right ]\\
              	&= - N \left( \texttt{KL}(q_{\phi_{\bm{\psi}}}(\bm{\psi}) \parallel p(\bm{\psi})) 
              	+ \texttt{KL}(q_{\phi_{\mathbf{t}}}(\mathbf{t}) \parallel p(\mathbf{t})) \right)\\
              	&\hphantom{= \;}+ \sum_{n=1}^{N} \mathbb{E}_{q_{\phi}} \left [\log p(\mathbf{x}_n | \mathbf{a}_n, \mathbf{t}, \bm{\psi}) \right ] - \texttt{KL}( q_{\phi_{\mathbf{a}}}(\mathbf{a}_n \,|\,\mathbf{x}_n) \parallel p(\mathbf{a}))\\
         	    &= - N \left( \texttt{KL}(q_{\phi_{\bm{\psi}}}(\bm{\psi}) \parallel p(\bm{\psi}))
         	    + \sum_{m=1}^{M} \texttt{KL}( q_{\phi_{\mathbf{t}}}(\mathbf{t}^m) \parallel p(\mathbf{t})) \right)\\
         	    &\hphantom{= \;}+ \sum_{n=1}^{N} 
                \left (\frac{1}{L} \sum_{l=1}^{L} \sum_{m=1}^{M} \log p(\mathbf{x}_n^m|\mathbf{a}_n^l, \mathbf{t}^{m,l}, \bm{\psi}^{l}) \right) 
                - \texttt{KL}( q_{\phi_{\mathbf{a}}}(\mathbf{a}_n \,|\,\mathbf{x}_n) \parallel p(\mathbf{a})).
\end{aligned}
\end{equation*}
\newpage

\subsection{Supplemental Results}

\begin{table}[ht]
\caption{Distance and correlation between actual and estimated branch lengths by EvoVGM\_JC69.}
\label{tab:jc_b}
\begin{center}
\begin{small}
\begin{sc}
\begin{tabular}{lccccccccc}
\toprule
{\(N \rightarrow\)} & \multicolumn{3}{c}{100} & \multicolumn{3}{c}{1000} & \multicolumn{3}{c}{5000} \\
{\(M\)} &  Dist &  Corr &  pval &  Dist &  Corr &  pval &  Dist &  Corr &  pval \\
\midrule
3 & 0.129 & 0.969 & 0.160 & 0.069 & 0.982 & 0.121 & 0.143 & 0.982 & 0.120 \\
4 & 0.166 & 0.938 & 0.062 & 0.065 & 0.997 & 0.003 & 0.079 & 0.997 & 0.003 \\
5 & 0.179 & 0.841 & 0.074 & 0.096 & 0.993 & 0.001 & 0.076 & 0.990 & 0.001 \\
\bottomrule
\end{tabular}
\end{sc}
\end{small}
\end{center}
\end{table}
\begin{table}[ht]
\caption{Distance and correlation between actual and estimated branch lengths by EvoVGM\_K80.}
\label{tab:k80_b}
\begin{center}
\begin{small}
\begin{sc}
\begin{tabular}{lccccccccc}
\toprule
{\(N \rightarrow\)} & \multicolumn{3}{c}{100} & \multicolumn{3}{c}{1000} & \multicolumn{3}{c}{5000} \\
{\(M\)} &  Dist &  Corr &  pval &  Dist &  Corr &  pval &  Dist &  Corr &  pval \\
\midrule
3 & 0.133 & 0.948 & 0.207 & 0.171 & 0.975 & 0.144 & 0.074 & 0.975 & 0.142 \\
4 & 0.184 & 0.855 & 0.145 & 0.093 & 0.996 & 0.004 & 0.049 & 0.992 & 0.008 \\
5 & 0.180 & 0.835 & 0.078 & 0.062 & 0.990 & 0.001 & 0.073 & 0.999 & 0.000 \\
\bottomrule
\end{tabular}
\end{sc}
\end{small}
\end{center}
\end{table}


\begin{figure*}[]
    \begin{subfigure}[]{.85\textwidth}
    \centering
    \includegraphics[width=\textwidth, trim={0 0.75cm 0 0}, clip]{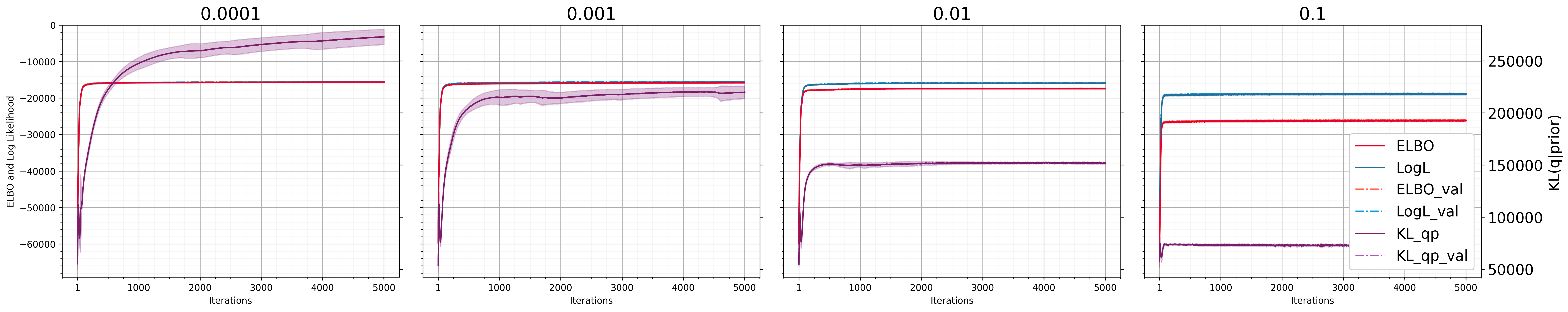}
    \end{subfigure}
    
    \begin{subfigure}[]{.85\textwidth}
    \centering
    \includegraphics[width=\textwidth, trim={0 0.85cm 0 0.85cm}, clip]{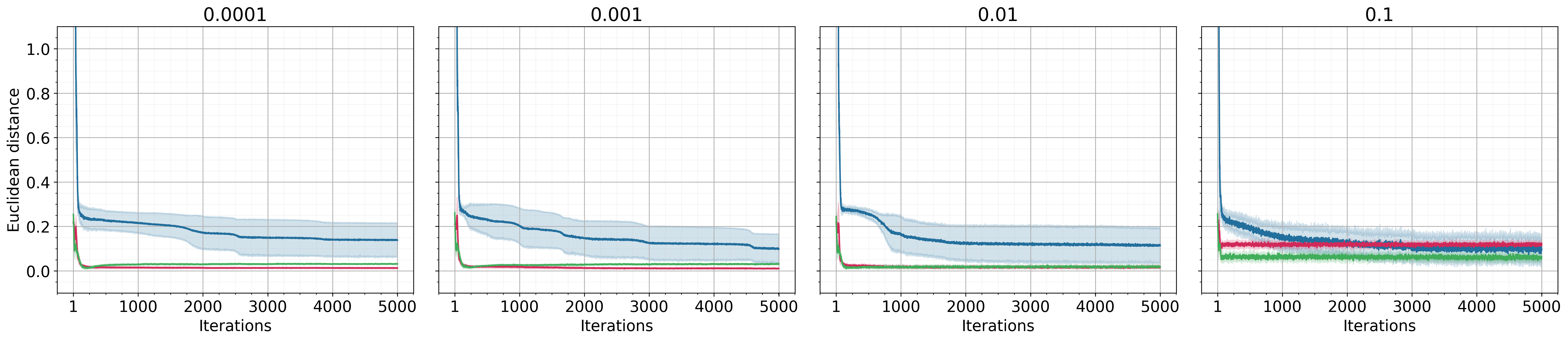}
    \end{subfigure}
    
    \begin{subfigure}[]{.85\textwidth}
    \centering
    \includegraphics[width=\textwidth, trim={0 0 0 0.85cm}, clip]{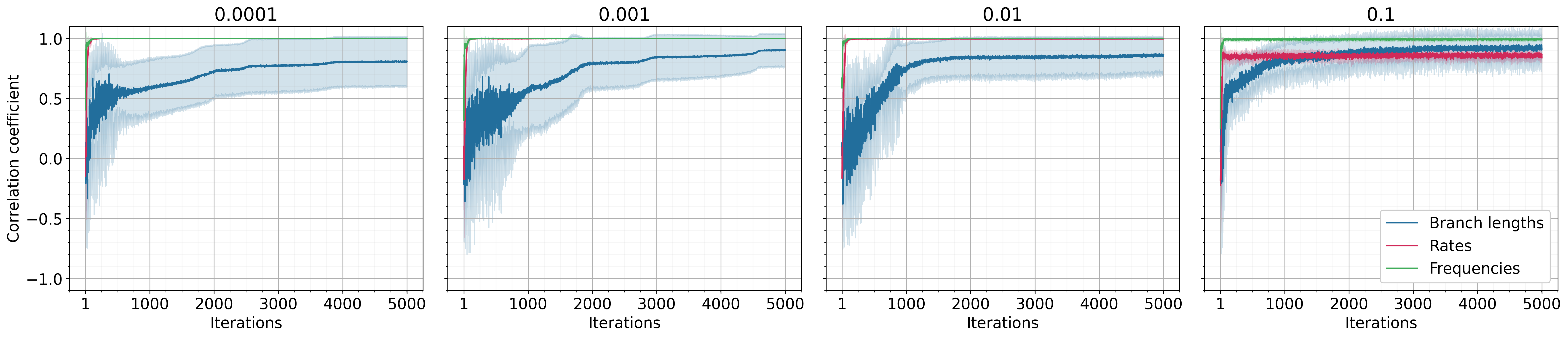}
    \end{subfigure}

    \caption{{\small Convergence and performance of EvoVGM\_GTR model for multiple settings of \(\alpha_{KL}\). The GTR substitution model was used to simulate training and validation alignments of five sequences with a length of 5000 bp. The estimates are computed and averaged from fitting and running the model ten times.}}
    \label{fig:gtr_kl}
\end{figure*}

\begin{figure*}[]
    \begin{subfigure}[]{0.85\textwidth}
    \centering
    \includegraphics[width=\textwidth, trim={0 0.8cm 0 0}, clip]{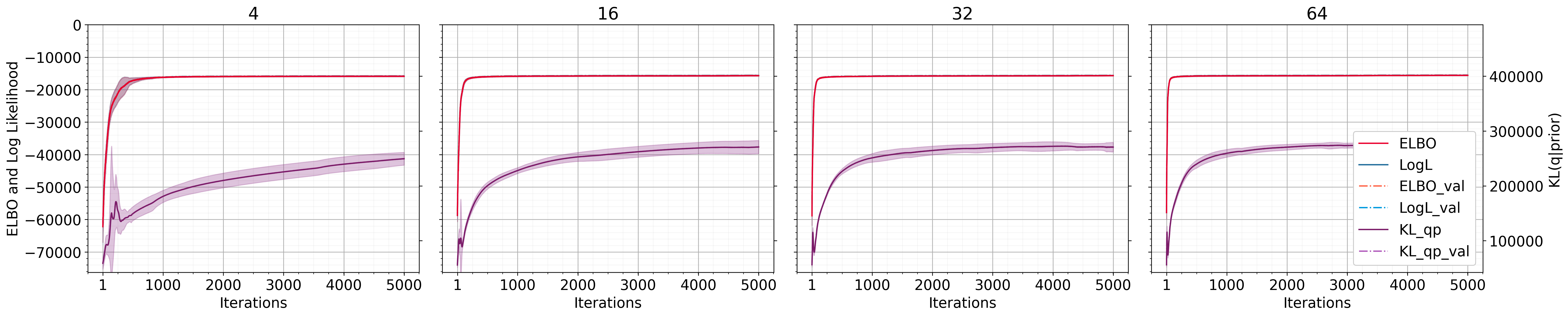}
    \end{subfigure}
    
    \begin{subfigure}[]{0.85\textwidth}
    \centering
    \includegraphics[width=\textwidth, trim={0 0.85cm 0 0.85cm}, clip]{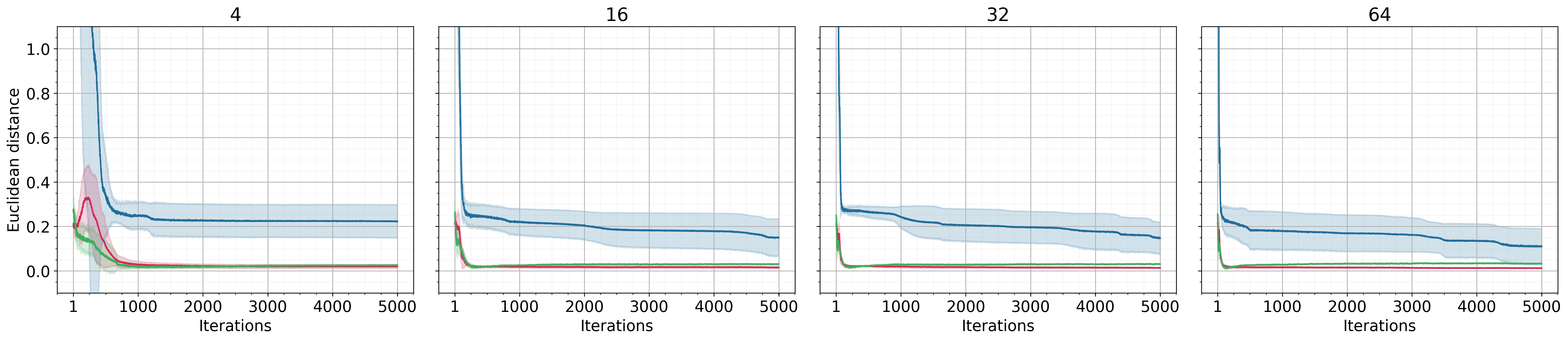}
    \end{subfigure}
    
    \begin{subfigure}[]{0.85\textwidth}
    \centering
    \includegraphics[width=\textwidth, trim={0 0 0 0.85cm}, clip]{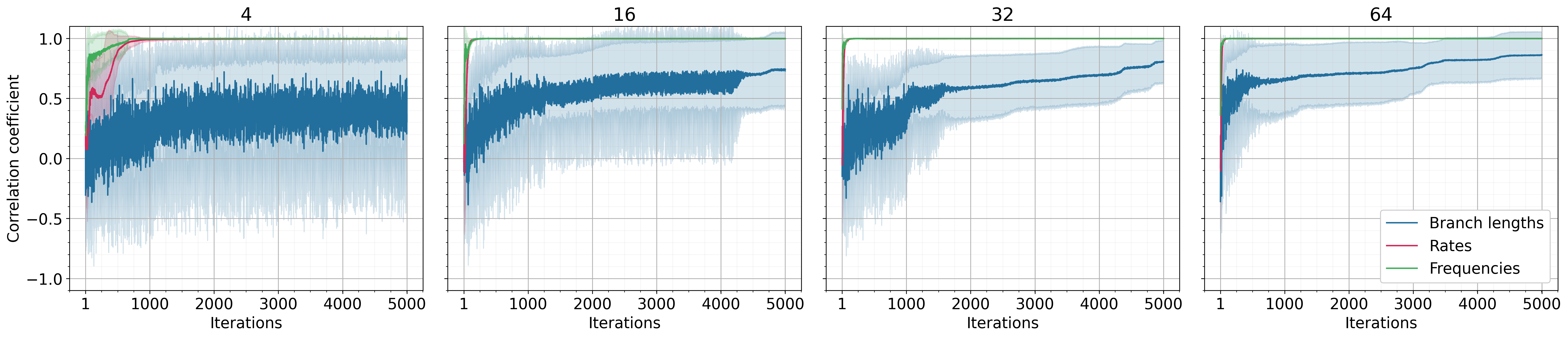}
    \end{subfigure}

    \caption{{\small Convergence and performance of EvoVGM\_GTR model for multiple settings of the hidden size of the neural networks of the variational encoders. The GTR substitution model was used to simulate training and validation alignments of five sequences with a length of 5000 bp. The estimates are computed and averaged from fitting and running the model ten times.}}
    \label{fig:gtr_hs}
\end{figure*}

\begin{figure*}[]
    \begin{subfigure}[]{0.85\textwidth}
    \centering
    \includegraphics[width=\textwidth, trim={0 0.8cm 0 0}, clip]{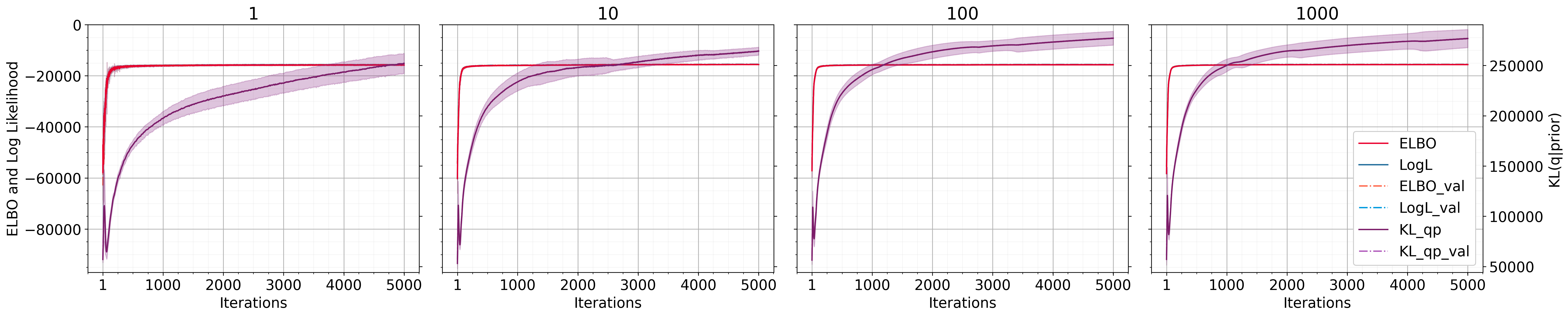}
    \end{subfigure}
    
    \begin{subfigure}[]{.85\textwidth}
    \centering
    \includegraphics[width=\textwidth, trim={0 0.85cm 0 0.85cm}, clip]{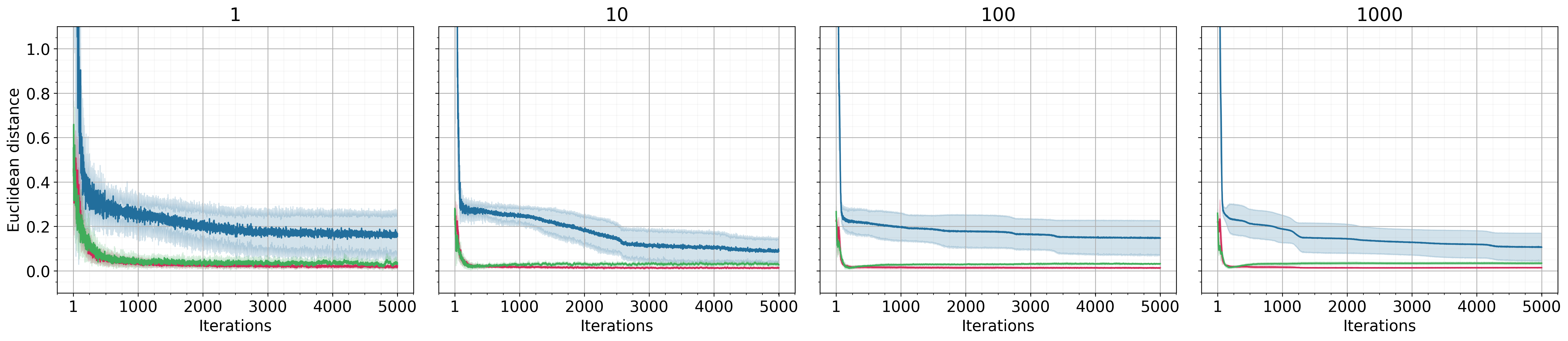}
    \end{subfigure}
    
    \begin{subfigure}[]{.85\textwidth}
    \centering
    \includegraphics[width=\textwidth, trim={0 0 0 0.85cm}, clip]{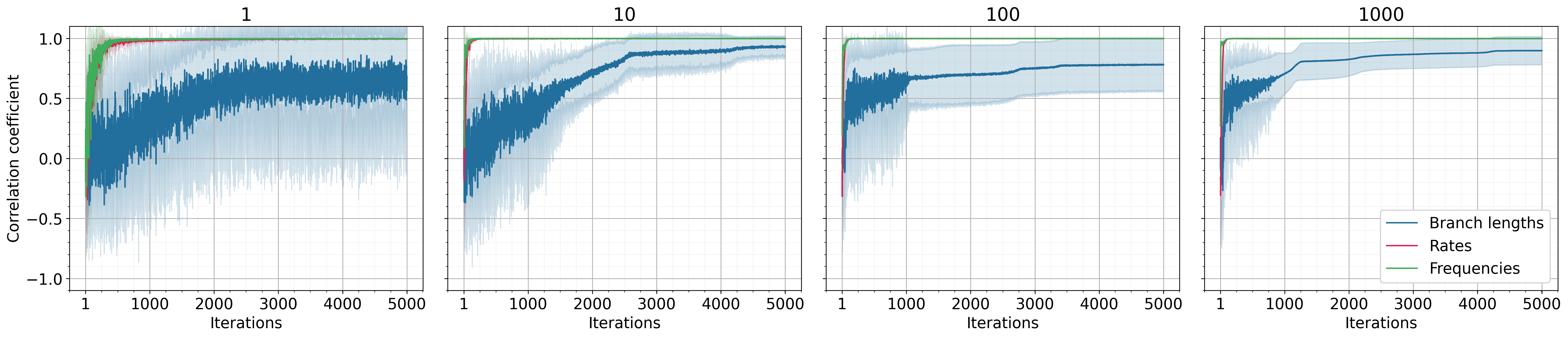}
    \end{subfigure}

    \caption{{\small Convergence and performance of EvoVGM\_GTR model for multiple settings of the sample size. The GTR substitution model was used to simulate training and validation alignments of five sequences with a length of 5000 bp. The estimates are computed and averaged from fitting and running the model ten times.}}
    \label{fig:gtr_ns}
\end{figure*}
\begin{figure*}[]
    \begin{subfigure}[]{.85\textwidth}
    \centering
    \includegraphics[width=\textwidth, trim={0 0.8cm 0 0}, clip]{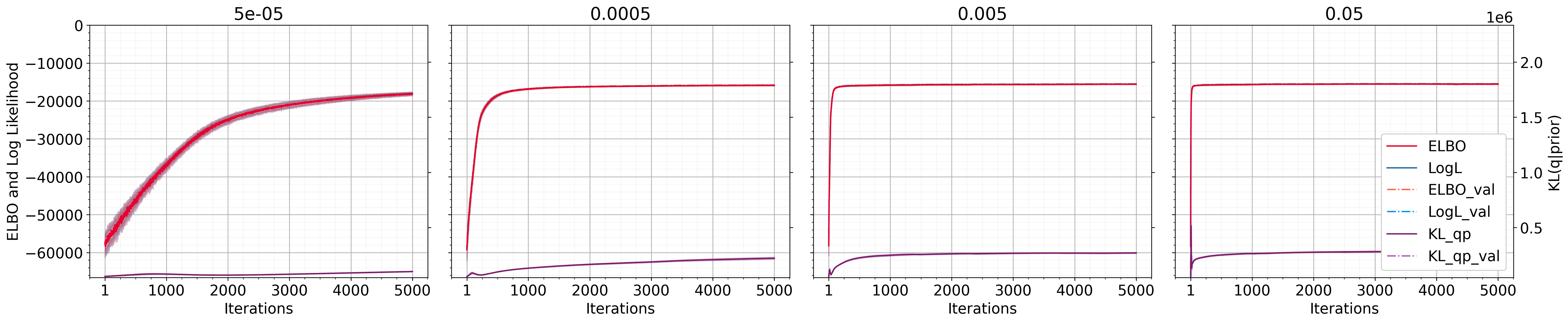}
    \end{subfigure}
    
    \begin{subfigure}[]{.85\textwidth}
    \centering
    \includegraphics[width=\textwidth, trim={0 0.85cm 0 0.85cm}, clip]{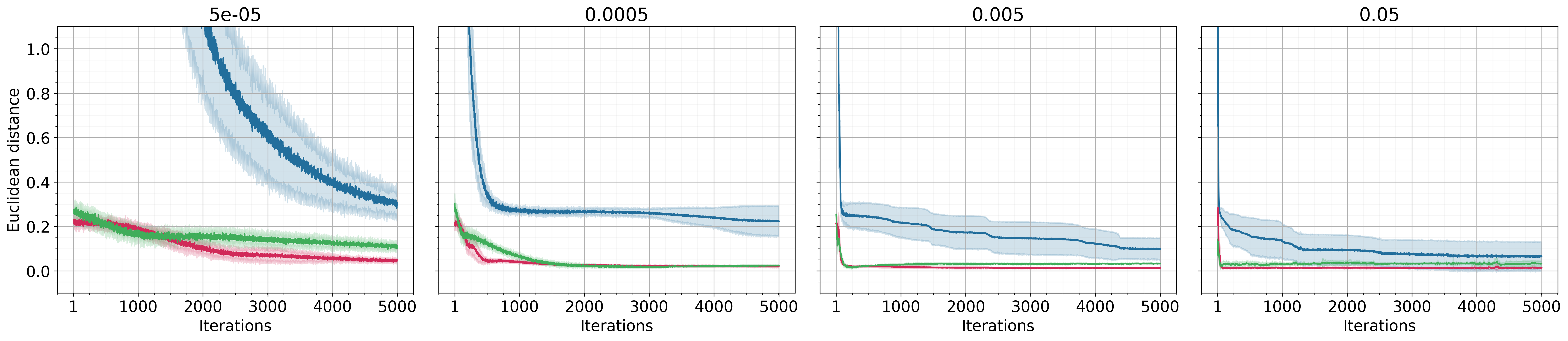}
    \end{subfigure}
    
    \begin{subfigure}[]{.85\textwidth}
    \centering
    \includegraphics[width=\textwidth, trim={0 0 0 0.85cm}, clip]{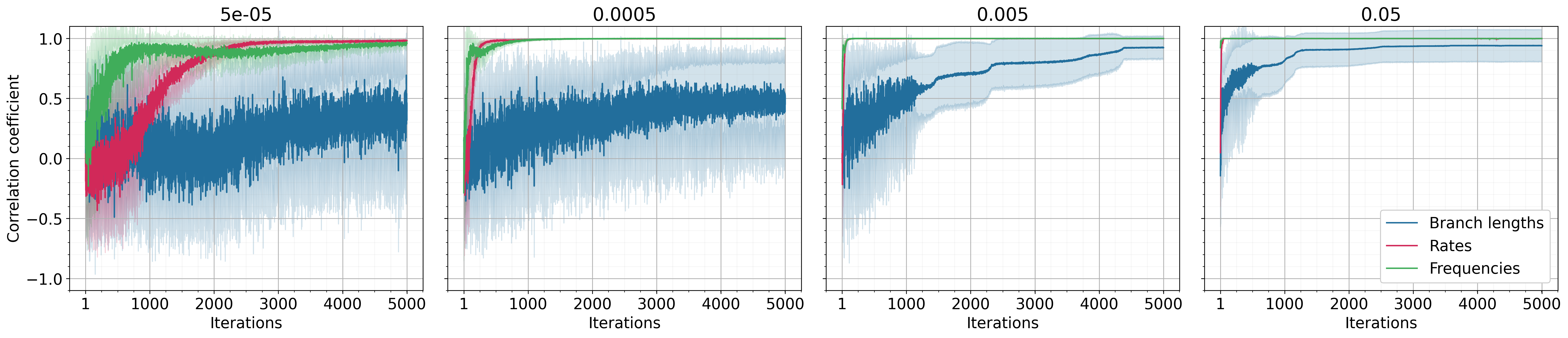}
    \end{subfigure}

    \caption{{\small Convergence and performance of EvoVGM\_GTR model for multiple settings of the learning rate. The GTR substitution model was used to simulate training and validation alignments of five sequences with a length of 5000 bp. The estimates are computed and averaged from fitting and running the model ten times.}}
    \label{fig:gtr_lr}
\end{figure*}

\end{document}